\documentclass[final,authoryear,3p,times]{elsarticle}

\usepackage{graphicx,color}
\usepackage{enumerate}
\usepackage[colorlinks]{hyperref}
\usepackage{natbib}
\usepackage{amsmath,amssymb,amsthm,bm}
\usepackage{multicol}
\usepackage[dvipsnames]{xcolor}
\usepackage{ulem}
\usepackage{tikz}
\usetikzlibrary{automata,positioning}
\usepackage{afterpage}
\usepackage{calc}
\usepackage{ifthen}
\usepackage{algorithm}
\usepackage{algpseudocode}
\usepackage{caption}
\usepackage{subcaption}
\usepackage{tabularx,ragged2e}
\usepackage{colortbl}
\usepackage{amssymb}
\usepackage[title]{appendix}
\usepackage{cancel}

\newcommand{\boldface}[1]{\boldsymbol{#1}}  
\newcommand{\bfa}{\boldface{a}}

\newcommand{\bfc}{\boldface{c}}

\newcommand{\bfu}{\boldface{u}}

\newcommand{\bfw}{\boldface{w}}

\newcommand{\bfz}{\boldface{z}}
\newcommand{\bfA}{\boldface{A}}
\newcommand{\bfB}{\boldface{B}}
\newcommand{\bfC}{\boldface{C}}

\newcommand{\bfE}{\boldface{E}}
\newcommand{\bfF}{\boldface{F}}

\newcommand{\bfH}{\boldface{H}}
\newcommand{\bfI}{\boldface{I}}

\newcommand{\bfP}{\boldface{P}}

\newcommand{\calA}{\mathcal{A}}
\newcommand{\calB}{\mathcal{B}}

\newcommand{\calD}{\mathcal{D}}

\newcommand{\calF}{\mathcal{F}}
\newcommand{\calG}{\mathcal{G}}

\newcommand{\calQ}{\mathcal{Q}}

\newcommand{\calU}{\mathcal{U}}

\newcommand{\calX}{\mathcal{X}}

\newcommand{\Rset}{\mathbb{R}}

\newlength{\boxwidth}
\def\dd{\;\!\mathrm{d}}

\def\btheorem{\begin{theorem}}
\def\etheorem{\end{theorem}}
\def\blemma{\begin{lemma}}
\def\elemma{\end{lemma}}
\def\bproposition{\begin{proposition}}
\def\eproposition{\end{proposition}}
\def\bcorollary{\begin{corollary}}
\def\ecorollary{\end{corollary}}
\def\bdefinition{\begin{definition}}
\def\edefinition{\end{definition}}
\def\bexample{\begin{example}}
\def\eexample{\end{example}}
\def\bremark{\begin{remark}}
\def\eremark{\end{remark}}

\newcommand{\eps}{\varepsilon}

\DeclareMathOperator{\tr}{tr}

\newcommand{\be}{\begin{equation}}
\newcommand{\ee}{\end{equation}}
\newcommand{\beq}{\begin{eqnarray}}
\newcommand{\eeq}{\end{eqnarray}}
\newcommand{\bem}{\begin{multline}}
\newcommand{\eem}{\end{multline}}
\newcommand{\ba}{\begin{align}}
\newcommand{\ea}{\end{align}}

\renewcommand{\figurename}{Figure }

\definecolor{Gray}{gray}{0.85}
\newcolumntype{v}{>{\arraybackslash\hsize=0.85\hsize}X}
\newcolumntype{u}{>{\arraybackslash\hsize=0.15\hsize}X}

\journal{}

\begin{document}

\begin{frontmatter}

\title{NN-EUCLID: deep-learning hyperelasticity without stress data}

\author[a]{Prakash Thakolkaran}
\author[a]{Akshay Joshi}
\author[a]{Yiwen Zheng}
\author[b]{Moritz Flaschel}
\author[b]{Laura De Lorenzis}
\author[a]{Siddhant Kumar\corref{cor1}}

\cortext[cor1]{Email: Sid.Kumar@tudelft.nl}

\address[a]{Department of Materials Science and Engineering, Delft University of Technology, 2628 CD Delft, The Netherlands}
\address[b]{Department of Mechanical and Process Engineering, ETH Zürich, 8092 Zürich, Switzerland}

\begin{abstract}

We propose a new approach for \textit{unsupervised}  learning of hyperelastic constitutive laws with physics-consistent deep neural networks. In contrast to supervised learning, which assumes the availability of stress-strain pairs, the approach only uses realistically measurable full-field displacement and global reaction force data, thus it lies within the scope of our recent framework for \textit{Efficient Unsupervised Constitutive Law Identification and Discovery} (EUCLID) and we denote it as NN-EUCLID. The absence of stress labels is compensated for by leveraging a physics-motivated loss function based on the conservation of linear momentum to guide the learning process. The constitutive model is based on input-convex neural networks, which are capable of learning a function that is convex with respect to its inputs.  By employing a specially designed neural network architecture, multiple physical and thermodynamic constraints for hyperelastic constitutive laws, such as material frame indifference, material stability, and stress-free reference configuration are automatically satisfied. We demonstrate the ability of the approach to accurately learn several hidden isotropic and anisotropic hyperelastic constitutive laws – including e.g., Mooney-Rivlin, Arruda-Boyce, Ogden, and Holzapfel models -- without using stress data. For anisotropic hyperelasticity, the unknown anisotropic fiber directions  are  automatically discovered jointly with the constitutive model. The   neural network-based constitutive models show good generalization capability beyond the strain states observed during training and are readily deployable in a general finite element framework for simulating complex mechanical boundary value problems with good accuracy.
\end{abstract}


\end{frontmatter}

\section{Introduction}
\label{sec:Introduction}

The merger of data-driven methods (often enabled or empowered by machine learning tools) and constitutive material modeling in solid mechanics is rapidly redefining the way we characterize and model complex material behavior. Traditionally, phenomenological models -- parameterized by a few material-specific constants -- are  calibrated by  iterative tuning to match the observations of simple mechanical tests, e.g., uni-/biaxial tension, torsion, and bending tests. Advances in experimental mechanics such as the advent of digital image correlation (DIC) and digital volume correlation (DVC) enabled more sophisticated approaches, such as finite element model updating (FEMU) \citep{Marwala2010} or the virtual fields method (VFM) \citep{Pierron_2012}, which rely upon the availability of full-field displacement data and exploit them for calibration of unknown parameters in a priori selected classical material models.

Modern data-driven methods aim to take a further step to overcome the limited expressive power  of  classical material models.  In the (material) model-free paradigm  proposed by \citet{kirchdoerfer_data-driven_2016} and first formulated for elasticity, a material's state of deformation is directly mapped to a stress state closest to the stress-strain pair available in the dataset (subject to physical compatibility constraints) \citep{kirchdoerfer_data-driven_2016,ibanez_data-driven_2017,Kirchdoerfer_dynamics_2017,conti_data-driven_2018,nguyen_data-driven_2018,eggersmann_model-free_2019,carrara_data-driven_2020,karapiperis_data-driven_2021}.  Alternative approaches keep the concept of a material model and surrogate it by learning an approximate mapping between the strains and stresses (referring again to the easiest case of elasticity) using, e.g., sparse regression with feature engineering \citep{flaschel_unsupervised_2021,flaschel_plasticity,joshi_Bayesian, Wang2021}, manifold learning methods and polynomial approximations \citep{ibanez_manifold_2018, ibanez_data-driven_2017,gonzalez_thermodynamically_2019}, Gaussian process regression \citep{rocha_onthefly_2021,fuhg_local_2022}, and artificial neural networks (NNs) \citep{ghaboussi_knowledgebased_1991,fernandez_anisotropic_2021,klein_polyconvex_2022,vlassis_sobolev_2021,kumar_inverse-designed_2020,bastek_inverting_2022,zheng_data_2021,Mozaffar2019,Bonatti2021,Vlassis2020,kumar_ml_mech_review,Asad2022,Liang2022}, with the list being roughly in order of decreasing physical interpretability and increasing approximation power. Further hybrid approaches use data to construct automatic corrections to existing models \citep{ibanez_hybrid_2019,gonzalez_learning_2019}. Physics-informed neural networks (PINNs) \citep{Raissi2019} have been used for inverse estimation of constitutive models or differential equations, but they are limited to either constitutive models of known form with unknown parameters or unknown constitutive models but for one-dimensional cases \citep{huang_learning_2020,tartakovsky_learning_2018,haghighat_deep_2020,Chen2021}. Consequently, they suffer from limited model expressive power  -- similar to parametric model calibration with classical FEMU and VFM techniques. 

A common bottleneck in most state-of-the-art data-driven methods, be them based on NNs or not, is the requirement of a large number of stress labels, i.e., in machine learning jargon, the methods are \textit{supervised}. In principle, stress labels can be computationally generated via multiscale homogenization techniques; however, this data generation approach relies upon modeling at the lower scale(s) (where model choice and calibration is even more challenging), is associated with a prohibitive computational cost, and its application is restricted to surrogate modeling and acceleration of multiscale simulations. Obtaining stress labels experimentally is a challenge too. Mechanical experiments from which stresses can be inferred, such as uni-/bi-axial tensile tests or bending tests, are too simple to sufficiently probe the high-dimensional stress-strain space and more complex experiments only provide boundary projections of stress tensors in form of force measurements.   The recent recognition of this issue motivated the development of the data-driven identification
method \citep{leygue_data-based_2018,dalemat_measuring_2019,Cameron2021}, which formulates the inverse problem associated
to the approach in \cite{kirchdoerfer_data-driven_2016}. In this light, breaking away from the supervised setting, i.e., reliance on stress labels, is the key to developing truly data-efficient and practically applicable data-driven constitutive models.

To address the aforementioned limitations of classical approaches and supervised data-driven methods, the authors recently proposed a novel framework denoted as \textit{Efficient Unsupervised Constitutive Law Identification and Discovery} (EUCLID) \citep{flaschel_unsupervised_2021,flaschel_plasticity,joshi_Bayesian}. The method is unsupervised, i.e., it requires no stress data but only global reaction forces and full-field displacement data  realistically obtainable through, e.g., DIC or DVC; it uses sparse regression over a large catalog of candidate functions to deliver interpretable constitutive laws given by parsimonious mathematical expressions. EUCLID departs from data-driven learning to achieve physics-driven learning, i.e., the need for stress labels is circumvented by enforcing that the full-field displacement data satisfy the conservation of linear momentum -- possibly along with additional physics-based requirements. We recently demonstrated the performance of EUCLID in isotropic hyperelasticity \citep{flaschel_unsupervised_2021}, rate-independent plasticity \citep{flaschel_plasticity}, anisotropic hyperelasticity and elastodynamics \citep{joshi_Bayesian}. In the latter contribution, we developed Bayesian-EUCLID, a probabilistic framework based on Bayesian learning which automatically discovers material models along with the related uncertainties.

Despite the advantages in terms of interpretability, models discovered by sparse regression are limited to the space spanned by the linear (or non-linear) combination of hand-crafted candidate functions in the feature library.   On the other hand NNs, being universal approximators \citep{Hornik1989}, greatly surpass sparse regression in terms of size of the model space. At the expense of interpretability, they can, in principle, represent any generic constitutive behavior -- with no limits on expressive power and no need to rely upon domain knowledge. However,  the approximation power comes along with a high risk for overfitting; this issue and the high degree of ill-posedness due to the lack of stress labels hindered the development of unsupervised NN-based constitutive models thus far.

In this work, within the scope of the EUCLID framework, we demonstrate the first case of unsupervised NNs-based learning of constitutive models with no stress data. In this first step, we focus on isotropic and anisotropic hyperelasticity. To represent the constitutive model, we use input convex neural networks (ICNN) -- recently developed by \citet{amos_input_2017} and applied to constitutive modeling  \citep{Asad2022,klein_polyconvex_2022,FUHG2022101446,HUANG2022104856}.  Their special architecture automatically guarantees material stability for the underlying strain energy density, material objectivity, and stress-free reference configuration. In the absence of stress labels, the NN is trained to a physics-constrained loss based on full-field displacement data satisfying conservation of linear momentum. The approach delivers unsupervised NN-based constitutive models that are physically admissible, can generalize well beyond the strain states observed in the full-field displacement data, and can automatically discover the unknown principal directions of anisotropy whenever applicable. We denote the approach as NN-EUCLID.

\section{Unsupervised deep learning of hyperelastic constitutive laws}\label{sec:Methods}

\subsection{Problem setting}\label{subsec:Problem}
Consider a specimen of an unknown hyperelastic material subjected to quasi-static mechanical deformation with two-dimensional reference domain $\Omega \in \mathbb{R}^2$. A complex specimen geometry (e.g., plate with a hole) is specifically chosen to generate heterogeneous and diverse strain states. The material is assumed to be homogeneous and possibly, anisotropic with unknown fiber orientations $\{\alpha_i\in[0,\pi):i=1,2,\dots\}$. Note that the fiber orientations have two-fold symmetry and, therefore, are constrained to $[0,\pi)$ instead of $[0,2\pi)$. We apply Dirichlet and Neumann conditions on the boundaries $\partial \Omega_u  \ \subseteq  \ \partial \Omega$ and $ \ \partial \Omega_t  \  =   \ \partial \Omega \setminus \partial \Omega_u$, respectively. Without losing generality, we only consider displacement-controlled conditions (i.e., Dirichlet boundary conditions) in the subsequent discussions, while noting that applied forces in load-controlled conditions are equivalent to reaction forces in displacement-controlled conditions. The observed data consists of $n_t$ snapshots of displacement measurements $\calU = \{\bfu^{a,t} \in \mathbb{R}^2 : a = 1, \dots, n_n; t=1,\dots,n_t\}$ at $n_n$  points with coordinates $\mathcal{X} =\{\boldsymbol{X}^a \in \Omega : a = 1,\dots,n_n \}$ in the reference domain. Additionally, for each snapshot, $n_\beta$ reaction forces $\{R^{\beta,t}: \beta=1,\dots,n_\beta; t=1,\dots,n_t\}$ are assumed to be measured (using, e.g., load cells) at some but not necessarily all the sets of Dirichlet boundary conditions. In the subsequent discussion, we drop the superscript $(\cdot)^t$ for the sake of brevity; however the numerical procedure applies to all snapshots independently.

\begin{figure}[t]
		\centering
		\includegraphics[width=1.0\textwidth]{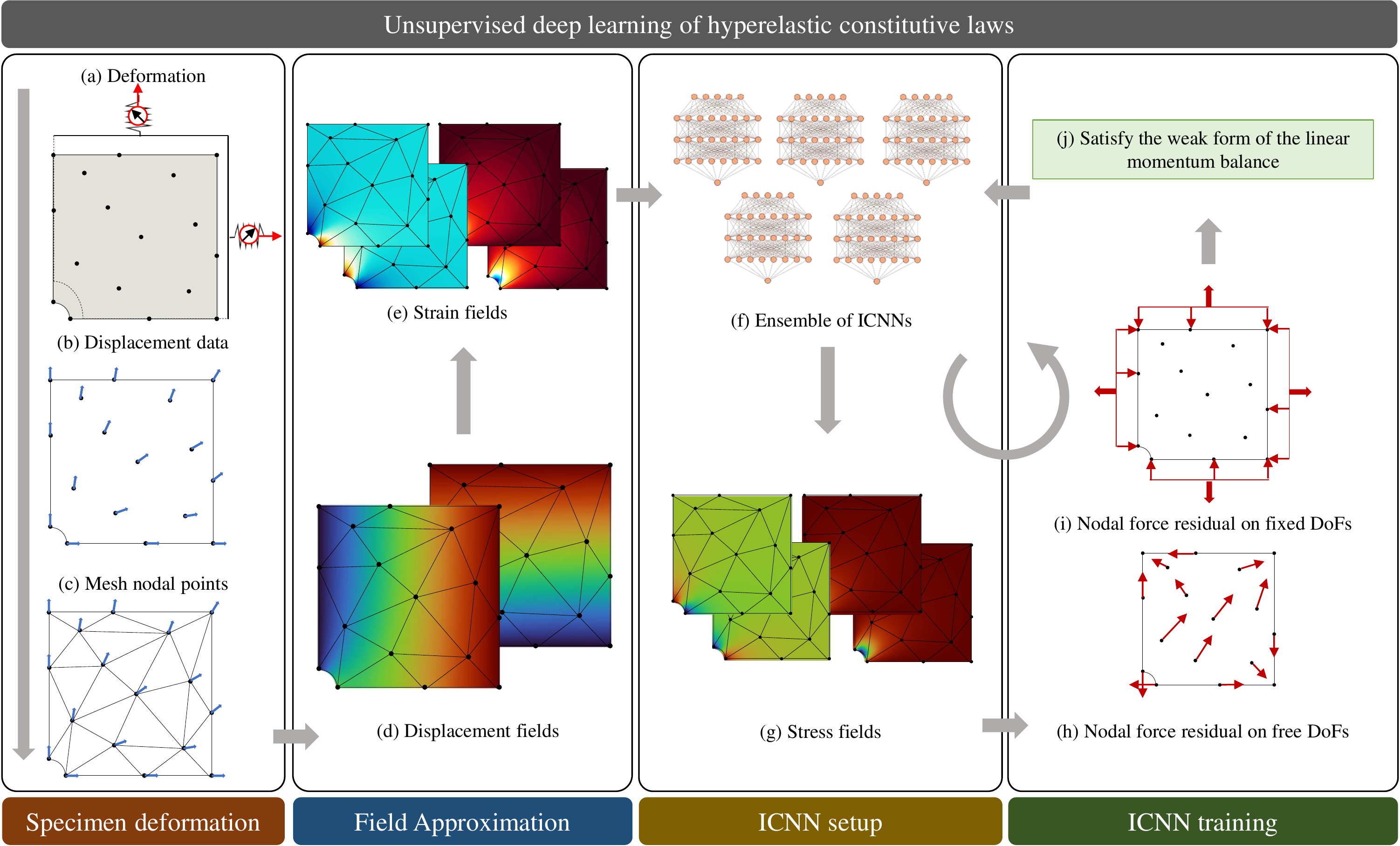}
		\caption{Schematic of the approach for unsupervised deep learning of hyperelastic constitutive models. (a,b) Point-wise displacements and reaction forces are recorded for a hyperelastic specimen under quasi-static deformation. Using the point-wise displacements and a finite element mesh of the domain (c), continuous displacement and strain fields are obtained (d,e). An ensemble of physics-consistent ICNN-based constitutive models (f) map the strain fields to stress fields (evaluated at the quadrature point of each element) (g). The stress fields are used to compute the internal and external nodal forces (h,i). Based on the weak form of the conservation of linear momentum, the residual forces are minimized (j) -- point-wise  for the free degrees of freedom (h) and aggregated  for the fixed degrees of freedom (under each set of Dirichlet constraint with a measured reaction force) (i). The optimization is carried out iteratively to train the parameters of the ICNN-based constitutive models (f).} \label{fig:methods_overview}
\end{figure}

With this limited data, the objective is to learn the underlying constitutive model governing the stress-strain response. While most modern DIC setups can easily track the displacement of thousands to millions of  points, it is impractical to have more than a few load cells attached to a specimen, i.e., $n_\beta\ll n_n$. In addition, since reaction forces are only boundary-aggregated projections of stress tensors, the lack of supervision, i.e., stress labels for constitutive model learning becomes apparent. Note that while the data contains two-dimensional displacements, the constitutive model that will be learnt is three-dimensional. The following sections describe the unsupervised learning approach using the aforementioned data. \figurename\ref{fig:methods_overview} illustrates a step-by-step schematic of the approach.

\subsection{Approximation of the displacement field from point-wise data}\label{subsec:Approx}

We mesh the points $\mathcal{X}$ in the reference domain using  linear triangular elements (with single quadrature point at the barycenter) to obtain the displacement field as
\be\label{disp-field-approx}
\boldsymbol{u}(\boldsymbol{X}) = \sum_{a=1}^{n_n}N^a(\boldsymbol{X})\ \boldsymbol{u}^{a}.
\ee
Here $N^a : \Omega \rightarrow \mathbb{R}$ is the shape function (corresponding to linear triangular elements) associated with the node at $\boldsymbol{X}^a$. The deformation gradient field is then approximated as
\begin{align}
\boldsymbol{F}(\boldsymbol{X}) = \boldsymbol{I} + \sum_{a=1}^{n_n}\boldsymbol{u}^{a}\otimes \nabla N^a(\boldsymbol{X}),
\end{align}where $\boldsymbol{I}$ is the identity matrix and $\nabla$ is the gradient operator with respect to the reference coordinates. 

\subsection{NN-based constitutive  model}\label{subsec:ConstModel}

A hyperelastic constitutive model is given by a strain energy density $W(\bfF)$ from which the first Piola-Kirchhoff stress $\bfP(\bfF)$ and the (incremental) tangent modulus $\mathbb{C}(\bfF)$ are derived as 
\be\label{eq:PK-general}
    \bfP(\bfF) = \frac{\partial W(\bfF)}{\partial \bfF}\qquad \text{and}\qquad \mathbb{C}(\bfF) = \frac{\partial \bfP(\bfF)}{\partial \bfF},\qquad \forall \quad  \bfF \in \text{GL}_+(3),
\ee
respectively, where $\text{GL}_+(3)$ denotes the set of all  invertible second-order tensors with positive determinant. Therefore, the objective reduces to learning $W(\bfF)$. However, any learned model for $W(\bfF)$ must also satisfy the following physical and thermodynamic constraints:
\begin{itemize}
  \item The stress must vanish in the case of no deformation (reference configuration), i.e.,
  \begin{align}
      \boldsymbol{P}(\boldsymbol{F}=\bfI)  = \boldsymbol{0}.
      \label{stress-free-condition}
  \end{align}
  \item The strain energy density must be objective, i.e.,
  \begin{align}
      W(\boldsymbol{R}\boldsymbol{F}) = W(\boldsymbol{F}), \ \ \  \forall\ \ \  \boldsymbol{F} \in \text{GL}_+(3), \ \ \boldsymbol{R} \in \text{SO}(3),
      \label{objectivity-condition}
  \end{align}
  where \text{SO}(3) denotes the 3D rotation group. 
  \item For material stability, $W(\boldsymbol{F})$ must satisfy quasiconvexity \citep{Morrey1952,Schroder2010}, i.e.,
  \be
    \int_\calB W(\bar\bfF + \nabla \bfw)\dd V \geq  W(\bar \bfF)\int_\calB \dd V \qquad \forall \qquad \calB\subset\Rset^3, \ \ \bar\bfF\in\text{GL}_{+}(3), \ \ \bfw\in C^\infty_0(\calB) \ \  \text{(i.e., $\bfw=0$ on $\partial\calB$)}.
  \ee
  However, enforcing quasiconvexity is often analytically/numerically intractable \citep{Kumar2019}. Since polyconvexity implies quasiconvexity, the quasiconvexity constraint is commonly relaxed to that of polyconvexity \citep{ball_convexity_1976,Schroder2010}. $W(\bfF)$ is polyconvex if and only if there exists a convex function $\mathcal{P}$ such that
  \begin{align}\label{eq:polyconvex}
      W(\boldsymbol{F}) = \mathcal{P}(\boldsymbol{F}, \text{Cof} \ \boldsymbol{F}, \det \boldsymbol{F}).
  \end{align}
\end{itemize}

To satisfy objectivity in \eqref{objectivity-condition}, a straightforward (and common) way of proceeding is to express the strain energy density as a function of the right Cauchy–Green deformation tensor $\bfC=\bfF^T\bfF$, or, equivalently, of the Green-Lagrange strain tensor $\bfE=(\bfF^T\bfF-\bfI)/2$, i.e., as $W(\bfE(\bfF))$, which represents a sufficient condition for objectivity. However, $\bfE$ is not necessarily convex in $\bfF$, hence this choice makes it difficult to enforce polyconvexity in $\bfF$ using \eqref{eq:polyconvex}. Following \citet{YangEtAl2017} and \citet{Asad2022}, we notice that the convexity of $W$ in $\bfE$ implies its local convexity in $\bfF$, which in turn ensures local material stability. Note that local convexity in $\bfF$ is preferrable to global convexity in $\bfF$, as the latter would preclude the non-uniqueness of the solution in situations where such non-uniqueness is physical (such as in buckling) and would prevent the satisfaction of the so-called growth condition, i.e. the condition of the strain energy density becoming infinite for $\det\bfF\rightarrow0^{+}$.

In light of the aforementioned constraints,  we consider an ansatz for the strain energy density of the form:
\be\label{eq:ansatz}
W(\bfF) \ =\  \underbrace{W^{\text{NN}}_{\calQ,\calA}\ (\bfE(\bfF))}_{\text{ICNN model}}
\ +  \underbrace{W^\text{0}}_{\substack{\text{Energy}\\\text{correction}}}
\ +\  \underbrace{\bfH:\bfE}_{\substack{\text{Stress}\\\text{correction}}} \ ,
\ee
which is adapted with modifications from \citet{Asad2022} and \citet{klein_polyconvex_2022}.
Here, $W^{\text{NN}}_{\calQ,\calA}$ represents an ICNN model \citep{amos_input_2017} with parameter sets $\calQ$ and $\calA$ (details described below), $W^0$ is a constant scalar, and $\bfH$ is a symmetric $3\times 3$ constant matrix. $W^0$ offsets the energy density such that it vanishes at zero deformation ($\bfF=\bfI$), i.e.,
\be\label{zero-energy}
W(\bfI) = 0 \qquad \implies \qquad W^0 = \ -\  \left.W^{\text{NN}}_{\calQ,\calA}\right|_{\bfF=\bfI}.
\ee
From \eqref{eq:PK-general}, the first Piola-Kirchhoff stress is obtained as
\be\label{eq:PK_ansatz}
\bfP(\bfF) = \frac{\partial W^{\text{NN}}_{\calQ,\calA}(\bfE(\bfF))}{\partial \bfF} +  \bfF\bfH.
\ee
Similar to $W^0$, $\bfH$ is set such that the stress vanishes at $\bfF=\bfI$ (see \eqref{stress-free-condition}), i.e.,
\be\label{H-matrix}
\bfP(\bfI) = \boldsymbol{0} \qquad \implies \qquad  \bfH \ = -\ \left.\frac{\partial W^{\text{NN}}_{\calQ,\calA}}{\partial \bfF}\right|_{\bfF=\bfI}.
\ee
The tangent modulus is derived from \eqref{eq:PK_ansatz} as 
\be\label{eq:tangent}
\mathbb{C}_{ijkl} = \frac{\partial P_{ij}(\bfF)}{\partial F_{kl}} = \frac{\partial^2 W_{\calQ,\calA}^\text{NN}(\bfE(\bfF))}{\partial F_{ij}\partial F_{kl}} + \delta_{ik}H_{lj},
\ee
where we use the Einstein summation convention over the subscripts and $\delta$ denotes the Kronecker delta. Note that $W^0$ and $\bfH$ in the energy and stress corrections, respectively, are updated in situ during each iterative step of the  neural network training (discussed later in Section \ref{sec:loss}), i.e., the corrections are not applied \textit{a posteriori}. Further description regarding the training of $W^{\text{NN}}_{\calQ,\calA}$ is provided in Algorithm \ref{alg:pseudocode}.

\begin{figure}[t]
		\centering
		\includegraphics[width=0.9\textwidth]{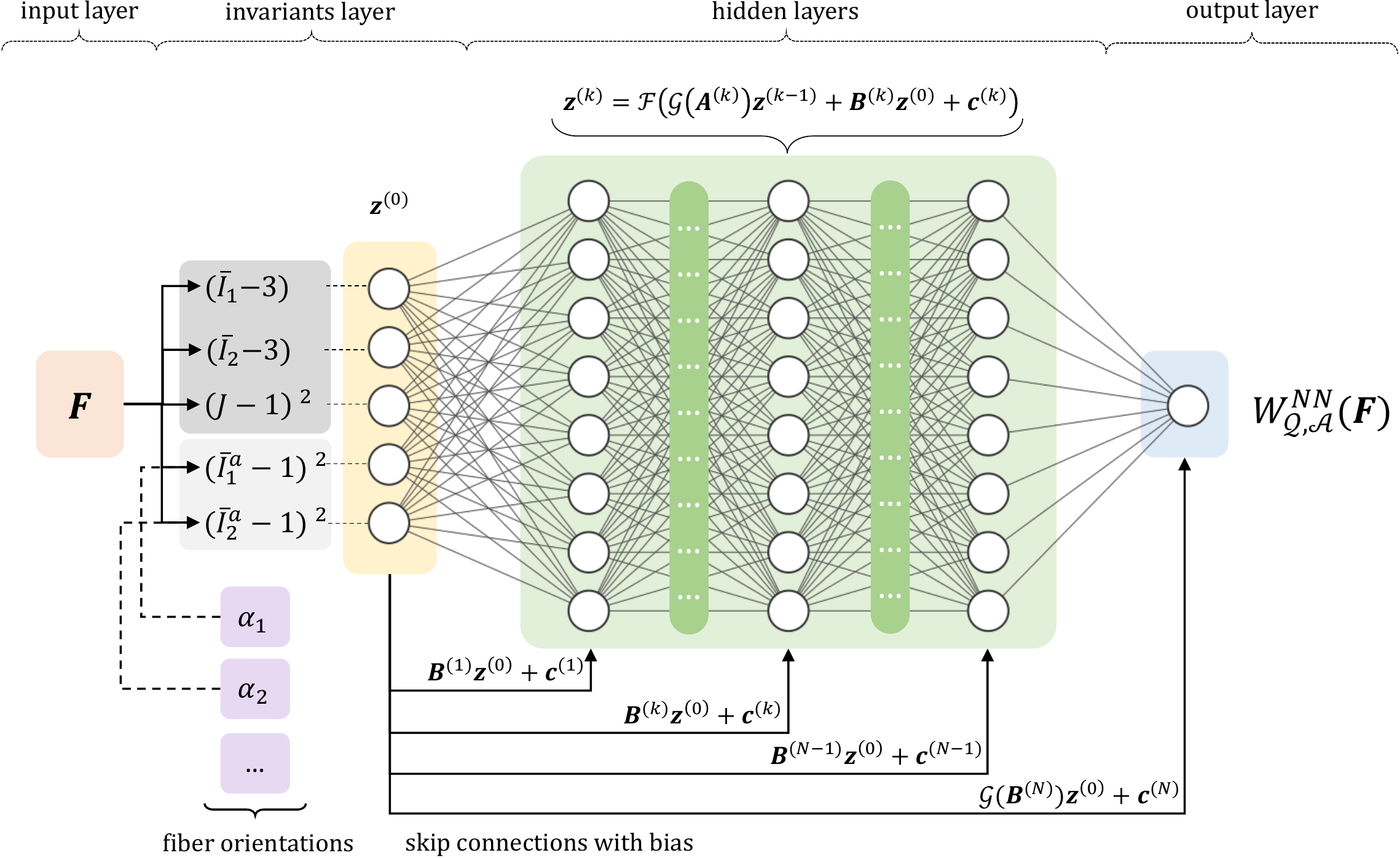}
		\caption{Schematic of the ICNN-based constitutive model for hyperelasticity.}\label{fig:icnn_schematic}
\end{figure}

$W^\text{NN}_{\calQ,\calA}$ -- enabled by the special ICNN architecture \citep{amos_input_2017} -- outputs a strain energy density that exhibits local convexity with respect to the input deformation gradient. Here, we present the architecture of $W^\text{NN}_{\calQ,\calA}$ followed by the description of each layer (see \figurename\ref{fig:icnn_schematic} for schematic).
\begin{subequations}
    \begin{align}
        \text{Input layer}: & \quad  \bfF\\
        \text{Invariants layer}: &\quad \bfz^{(0)} = \left[\tilde I_1-3,\ \tilde I_2-3, \ (J-1)^2,\  (\tilde I_{a_1}-1)^2,\ (\tilde I_{a_2}-1)^2,\ \dots\right]^T \\  
        &\quad \text{with trainable parameters:}\ \alpha_1 \in [0,\pi),\  \alpha_2 \in [0,\pi),\ \dots, \notag\\
        \text{First hidden layer of size $d_1$ }: & \quad \bfz^{(1)} = \calF\left( \calG\left(\bfA^{(1)}\right) \bfz^{(0)} + \bfB^{(1)} \bfz^{(0)} + \bfc^{(1)} \right)\\
        &\quad \text{with trainable parameters:}\ 
        \bfA^{(1)}\in\Rset^{d_1\times\left|\bfz^{(0)}\right|},\  
        \bfB^{(1)}\in\Rset^{d_1\times\left|\bfz^{(0)}\right|},\ 
        \bfc^{(1)}\in\Rset^{d_1}, \notag\\
        \vdots\qquad \qquad & \notag\\
        \text{$k^\text{th}$ hidden layer of size $d_k$ }: & \quad \bfz^{(k)} = \calF\bigg( \underbrace{\calG\left(\bfA^{(k)}\right) \bfz^{(k-1)}}_{\text{convex-linear}} + \underbrace{\bfB^{(k)} \bfz^{(0)}}_{\text{skip-connect}} + \underbrace{\bfc^{(k)}}_{\text{bias}} \bigg)\\
        & \quad \text{with trainable parameters:}\ 
        \bfA^{(k)}\in\Rset^{d_k\times d_{k-1}},\  
        \bfB^{(k)}\in\Rset^{d_k\times \left|\bfz^{(0)}\right|},\ 
        \bfc^{(k)}\in\Rset^{d_k}, \notag\\
        \vdots\qquad \qquad & \notag\\
        \text{Output ($N^\text{th}$) layer of size $d_N=1$}: & \quad W^\text{NN}_{\calQ,\calA} = \bfz^{(N)} =  \calG\left(\bfA^{(N)}\right) \bfz^{(N-1)} + \calG\left(\bfB^{(N)}\right) \bfz^{(0)} + \bfc^{(N)} \\
        & \quad \text{with trainable parameters:}\ 
        \bfA^{(N)}\in\Rset^{d_N\times d_{N-1}},\  
        \bfB^{(N)}\in\Rset^{d_N\times \left|\bfz^{(0)}\right|},\ 
        \bfc^{(N)}\in\Rset^{d_N}. \notag
    \end{align}
\end{subequations} 
$\calQ=\{\bfA^{(k)},\bfB^{(k)},\bfc^{(k)}: k=1,\dots,N\}$ and $\calA=\{\alpha_i:i=1,2,\dots\}$ denote the set of all the trainable ICNN weights and unknown fiber orientations, respectively.
The invariants layer receives the deformation gradient $\bfF$ from the input layer and  computes strain invariants  as
\be
    \begin{aligned}
        \text{volumetric invariant}: \qquad & J=\det(\bfF)=I_3^{1/2}\\
        \text{isotropic deviatoric invariants}: \qquad & \tilde I_1=J^{-2/3}I_1, \quad \tilde I_2=J^{-4/3}I_2,\\
        \text{anisotropic deviatoric invariants}: \qquad & \tilde I^a_1 =  J^{-2/3}\left( \bfa^{(1)} \cdot \bfC \bfa^{(1)}\right), \quad \tilde I^a_2 =  J^{-2/3}\left( \bfa^{(2)} \cdot \bfC \bfa^{(2)}\right), \quad \dots \quad,
    \end{aligned}
\ee
where 
\be
    I_1=\tr(\bfC),\qquad I_2 = \frac{1}{2}\left[\tr(\bfC)^2-\tr(\bfC^2)\right], \qquad I_3 = \det(\bfC)
\ee
are the principal invariants of  $\bfC$ (equivalently, $\bfE=(\bfC-\bfI)/2$). Here, we assume a deviatoric-volumetric split for modeling compressible hyperelasticity. For an anisotropic hyperelastic material with fiber orientations $\{\alpha_i:i=1,2,\dots\}$, $\bfa^{(i)}=(\cos \alpha_i,\sin\alpha_i,0)^T$ denotes the $i^\text{th}$ fiber direction with third component being normal to the plane of $\Omega$. The fiber orientations in $\calA$ are unknown and, therefore, treated as trainable parameters of the ICNN model. The invariants are further shifted by  appropriate scalar values -- in some cases, also squared -- such that both their value and the respective derivatives with respect to $\bfF$  are zero in the case of no deformation, i.e., at $\bfF=\bfI$.

The strain invariants -- now vectorized as $\bfz^{(0)}$ -- are sequentially transformed through $k=1,\dots,(N-1)$ hidden layers. In any $k^\text{th}$ layer of size $d_k>0$, the output of the previous layer ($\bfz^{(k-1)}$) undergoes a \textit{convex linear transformation} via $\calG\left(\bfA^{(k)}\right)$, where $\bfA^{(k)}\in\Rset^{d_k\times d_{k-1}}$ is a trainable weight matrix and $\calG$ denotes a  nonlinear \textit{non-negative} function applied element-wise. The layer also includes a \textit{skip connection} that adds the output of the invariants layer, i.e., $\bfz^{(0)}$, after linear transformation via trainable weight matrix $\bfB^{(k)}\in\Rset^{d_k\times \left|\bfz^{(0)}\right|}$ and bias vector $\bfc^{(k)}\in\Rset^{d_k}$. Finally, the sum of the convex linear transformation and the skip connection are passed through a nonlinear activation function $\calF$ that is \textit{convex} and \textit{non-decreasing} and acts element-wise on the input. The output/$N^\text{th}$ layer consists of summing the convex linear transforms of both the output of the penultimate layer ($\bfz^{(N-1)}$) and the invariants layer ($\bfz^{(0)}$); however, the output does not pass through the activation function $\calF$. 
    
The aforementioned architectural choices are motivated by the following reasons. Since $W^\text{NN}_{\calQ,\calA}$ only depends on the strain invariants, which are objective, the objectivity constraint \eqref{objectivity-condition} is identically satisfied. In contrast to a classical feed-forward neural network, the combination of convex linear transformations enabled via non-negative  $\calG$ in the hidden layers (non-negative weighted sums of convex functions are convex) and non-decreasing convex activation function $\calF$ ensures that the ICNN output  is convex in the strain invariants (see \cite{boyd_convex_2004,amos_input_2017} for proofs) and consequently,  the local material stability ensured by the overall strain energy density in \eqref{eq:ansatz}.
Without loss of generality, $\calG$ and $\calF$ are chosen based on the \textit{softplus} function \citep{Asad2022} as
\be\label{activations}
   \calG(x) = c_\calG\underbrace{\log(1+e^x)}_{\text{softplus}}  \qquad \text{and} \qquad \calF(x) = c_\calF\underbrace{\left(\log(1+e^x)\right)^2}_{\text{squared-softplus}},
\ee    
respectively, for any $x\in\Rset$; where $c_\calG>0$ and $c_\calF>0$ are hyperparameters. The $C^\infty$ order of continuity of the softplus function ensures that the strain energy density and its derivatives -- stress and tangent -- are sufficiently smooth and continuous.\footnote{Activation functions such as $\text{ReLU}(\cdot)=\max(\cdot,0)$ and its variants, while common in deep learning, are not recommended in this context as they are not twice differentiable.}  The squaring of the softplus function in $\calF$ mitigates the problem of vanishing second derivative of the softplus function. This not only enables computing the tangent of the strain energy density (see \eqref{eq:tangent}), but also prevents vanishing gradients for backpropagation-based NN training \citep{Goodfellow-et-al-2016}.\footnote{Since the loss function is based on the first derivative of the NN, training requires computing the second derivative for optimization of the NN parameters; see Section~\ref{sec:loss} and \eqref{loss}.} The skip connections are added to avoid vanishing gradients, overfitting, and accuracy saturation (or even degradation) as the NN architecture becomes deeper (with increasing number of hidden layers) \citep{He2016}. 

\subsection{Unsupervised learning of constitutive models}\label{sec:loss}

In the absence of energy density/stress labels, we use the fact that the observed data must satisfy the conservation of linear momentum to guide the learning process \citep{flaschel_unsupervised_2021} and estimate the model parameters $\calQ$ and $\calA$ of the ICNN.
In the case of negligible body forces and quasi-static loading (i.e., negligible inertia), the weak form of the linear momentum balance in the reference domain $\Omega$ is given by
\be\label{weak-form}
    \int_{\Omega} \boldsymbol{P}:\nabla\boldsymbol{v}\ \dd V - \int_{\partial\Omega_t} \hat{\boldsymbol{t}}\cdot \boldsymbol{v}\ \dd S = 0 \ \ \ \forall \ \  \text{admissible} \ \ \boldsymbol{v},
\ee
where $\hat{\boldsymbol{t}}$ is the surface traction acting on $\partial \Omega_t$ and $\boldsymbol{v}$ is an admissible test function that is sufficiently regular and vanishes on the Dirichlet boundary $\partial\Omega_u$. Note that $\hat{\boldsymbol{t}}=\boldsymbol{0}$ in the context of purely displacement-controlled testing. We use the weak form of the linear momentum balance instead of the strong form because the latter requires double spatial derivatives, which are more sensitive to noise in the data. For the scope of this work, we consider a plane strain assumption and note that the  method may be extended to the three-dimensional case with, e.g., Digital Volume Correlation data. While a plane stress assumption is more appropriate for samples with small out-of-plane thickness and would require a reformulation of the proposed method, for the purpose of developing and testing the method and given that we consider artificial data here, the plane strain assumption is just as good as any other -- provided it is consistent with the way the benchmark data are generated.

Let $\calD=\{(a,i):a=1,\dots,n_n;i=1,2\}$ denote the set of degrees of freedom  such that the displacement of the $a^\text{th}$ node in the $i^\text{th}$ direction, i.e., $u^a_i$, is included in the observed data. We further divide all the degrees of freedom into the following mutually exclusive subsets of $\calD$,
\begin{itemize}
    \item $\calD^\text{free}$: degrees of freedom not subjected to Dirichlet constraints, i.e., $u^a_i$ is not fixed;
    \item $\calD^\text{fix}_\beta$ (with $\beta=1,\dots,n_\beta$): degrees of freedom under Dirichlet constraints that collectively contribute to the observed reaction force  $R^\beta$.
\end{itemize}

Similar to the displacement field approximation \eqref{disp-field-approx}, we approximate the test function as
\be
    \boldsymbol{v}(\boldsymbol{X}) = \sum_{a=1}^{n_n}N^a(\boldsymbol{X})\boldsymbol{v}^a, \qquad \text{with} \qquad v_i^a = 0 \quad \forall\quad  (a,i) \in\bigcup_{\beta=1}^{n_\beta} \mathcal{D}^\text{fix}_\beta
\ee
for admissibility. Subsequently, the momentum balance in \eqref{weak-form} reduces to
\be\label{reducedWeakForm}
    \sum_{a=1}^{n_n}{v}^a_i f^a_i = 0, \qquad \text{where}\qquad  f^a_i = \underbrace{\int_{\Omega} {P}_{ij}\ \nabla_j N^a \ \dd V}_{\text{internal force}} - \underbrace{\int_{\partial\Omega_t} \hat{{t}}_i N^a \dd S}_{\text{external force}}.
\ee
$f^a_i$ can be interpreted as the difference between the internal force arising from the constitutive model and the external force due to the applied tractions on Neumann boundaries.
Note that the integrals in \eqref{reducedWeakForm} are computed using numerical quadrature defined by the mesh over the point set $\calX$. 

Since the test functions are arbitrary, \eqref{reducedWeakForm} must be satisfied at each free degree of freedom (where $\nu^a_i$ does not vanish) independently, i.e.,
\be\label{free-constraints}
f^a_i = 0 \qquad \forall \quad (a,i)\in \calD^\text{free}.
\ee
However, at the fixed degrees of freedom, the  internal and external forces are balanced by the reaction force from the Dirichlet constraints. Point-wise reaction forces cannot be measured experimentally, and are hence assumed to be unavailable. Instead, only global reaction forces along the boundary segments are known. Therefore, the aggregated force balance for each measured  reaction force  yields
\be\label{fix-constraints}
\sum_{(a,i)\in\calD^\text{fix}_\beta} f^a_i = R^\beta \qquad \forall \qquad \beta=1,\dots,n_\beta,
\ee
where the summation is performed over the point-wise forces across all the degrees of freedom in the $\beta^\text{th}$ Dirichlet constraint, i.e., $\calD^\text{fix}_\beta$. Recall that $(\cdot)^t$ was dropped for the sake of brevity and the above force balance constraints must be satisfied for all the data snapshots at $t=1,\dots,n_t$.

The objective now reduces to learning the constitutive model $W(\bfF)$ in \eqref{eq:ansatz} (equivalently, $\bfP(\bfF)$ in \eqref{eq:PK_ansatz}),  parameterized by the ICNN weights $\calQ$ and anisotropic fiber orientations $\calA$, such that the full-field displacement and reaction force data satisfy the purely physics-based constraints \eqref{free-constraints} and \eqref{fix-constraints}. We pose this inverse problem as the minimization of a loss function based on the force balance residuals given by
\be\label{loss}
\calQ, \calA \leftarrow \arg \min_{\calQ,\calA} 
\sum_{t=1}^{n_t}\Bigg[\ 
\underbrace{
\sum_{(a,i)\in\calD^\text{free}} \left(f^{a,t}_i\right)^2 
}_{\text{free degrees of freedom}}
\ +\ \ 
\underbrace{
\sum_{\beta=1}^{n_\beta}\Big(R^{\beta,t} - \sum_{(a,i)\in\calD^\text{fix}_\beta} f^{a,t}_i\Big)^2
}_{\text{fixed degrees of freedom}}
\ \Bigg].
\ee

The optimization problem in \eqref{loss} is solved via gradient-based minimization (see \ref{sec:implementation-details} for implementation details). Since each transformation -- starting from the observed displacement and reaction force data to the constitutive model and the physics-constrained loss -- is differentiable, the gradient of the objective function with respect to the trainable parameters in $\calQ$ and $\calA$ is easily computed. This is enabled by \textit{automatic differentiation} \citep{Baydin} -- wherein each mathematical operation is tracked in a  computational graph and later the gradients are computed via the chain rule for differentiation.

The inverse problem in \eqref{loss} is highly ill-posed  as it admits multiple solutions due to two reasons. \textit{(i)} While the ICNN output is convex with respect to the invariants of $\bfC$, the loss function in \eqref{loss} is highly non-convex with respect to the trainable parameters of the ICNN. \textit{(ii)} The absence of strain energy density or stress labels and thus the training with  indirect labels in the form of the physics-constrained loss implies that several different models can likely explain the limited observed data. For these reasons, the optimization problem in \eqref{loss} admits several local minima and the final solution is sensitive to the initial state of the trainable parameters, i.e., $\calQ$ and $\calA$. To address this issue, we consider an ensemble of ICNNs -- each trained independently with different initial states chosen randomly; we only accept those models which have distinctively low loss values and reject those for which the loss function gets trapped in bad minima with high loss values  (see \ref{sec:implementation-details} for implementation details). In Section~\ref{sec:Results}, we show that despite the multiplicity in the solution/model space,  the constitutive responses of the accepted models are accurate and consistent with each other and the proposed approach is robust to this ill-posedness.

\section{Numerical benchmarks}
\subsection{Data generation}
\begin{figure}[t]
		\centering
		\includegraphics[width=1.0\textwidth]{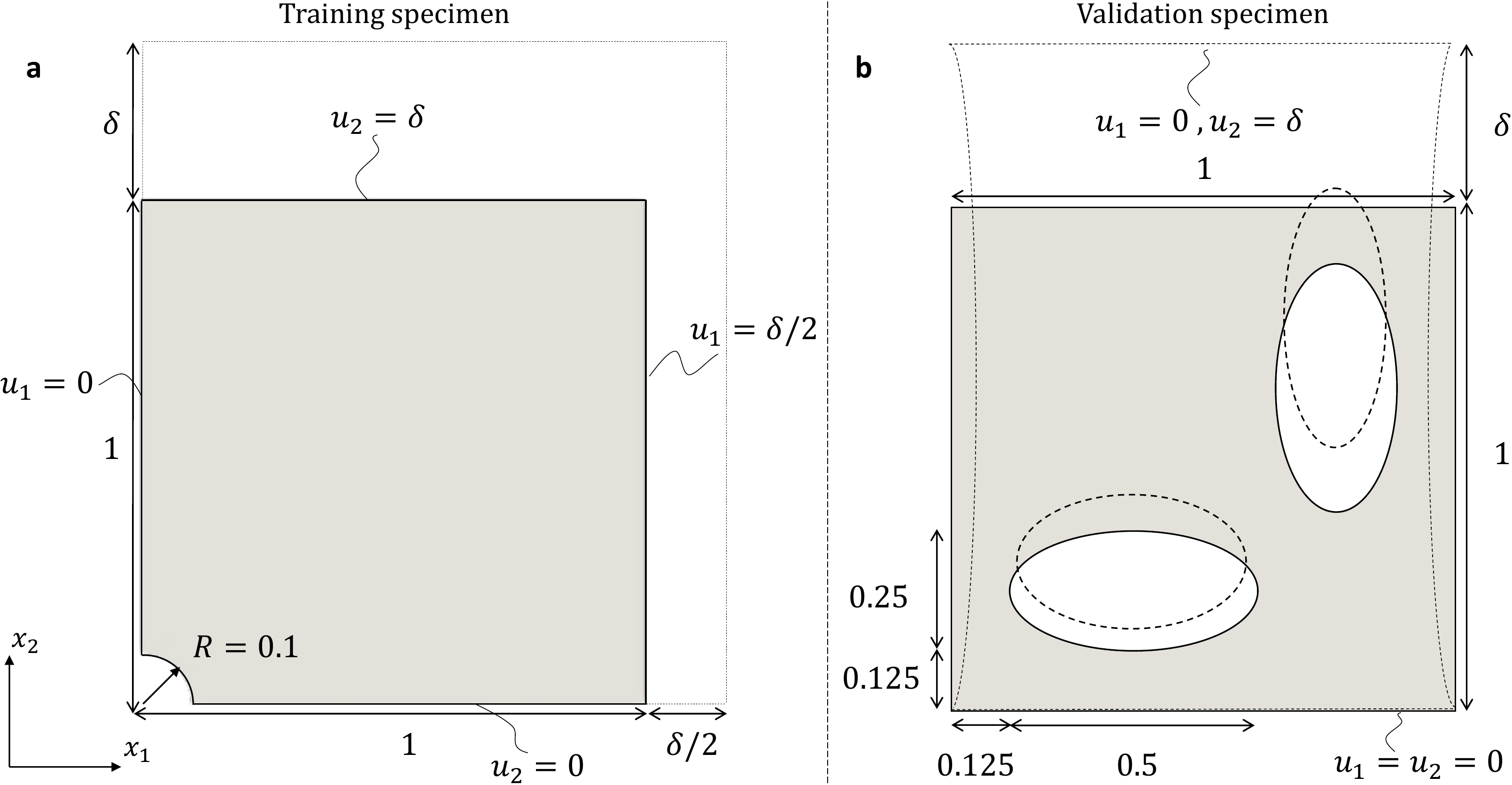}
		\caption{(a) \textit{Training specimen}: geometry and boundary conditions of a square plate with a hole in the bottom-left corner, subjected to displacement-controlled asymmetric biaxial tension. The resulting data (full-field displacements and reaction forces) including noise are used to train the ICNN-based constitutive models. (b) \textit{Validation specimen}: geometry and boundary conditions of a square plate with two asymmetric elliptical holes, subjected to displacement-controlled uniaxial tension. The specimen is only used for validation of the ICNN-based constitutive models;  no related data are used during the training stage. All lengths and displacements are normalized with respect to the side length of the undeformed specimen.}\label{fig:geometries}
\end{figure}

Adapting the benchmarks of \citet{flaschel_unsupervised_2021}, we emulate a DIC experiment by using the finite element method (FEM) to simulate the behavior of a hyperelastic square plate with a hole (shown in \figurename \ref{fig:geometries}a) under plane-strain conditions. The specimen undergoes displacement-controlled asymmetric biaxial tension with symmetry boundary conditions on the bottom and left boundaries and loading parameter $\delta$.  We employ linear triangular elements and the nodal displacements and reaction forces (horizontal force on the left and right boundaries, vertical force on the upper and lower boundaries) are recorded for $n_t$ load steps. Compared to simple uni-/biaxial tension or torsion tests, the combination of this geometry and loading generates sufficiently diverse and heterogeneous strain states to train a generalizable constitutive model with just a single experiment (see Section \ref{sec:Results} and \figurename\ref{fig:invariants-space_AB_IH}). Henceforth, the specimen is referred to as the \textit{training specimen}.

We generate the synthetic data with FEM using the following instances of well-known physical and phenomenological material models.
\begin{enumerate}
    \item Neo-Hookean (NH) model:
            \begin{equation}\label{eq:NH2}
                W(\bfF) = 0.5 (\Tilde{I}_1-3) + 1.5 (J-1)^2.
            \end{equation}
    \item Isihara (IH) model \citep{isihara_statistical_1951}:
            \begin{equation}\label{eq:IH}
                W(\bfF) = 0.5 (\Tilde{I}_1-3) +  (\Tilde{I}_2-3) +  (\Tilde{I}_1-3)^2 + 1.5 (J-1)^2.
            \end{equation}
    \item Haines-Wilson (HW) model \citep{haines_strain-energy_1979}:
            \begin{equation}\label{eq:HW}
                W(\bfF) = 0.5 (\Tilde{I}_1-3) +  (\Tilde{I}_2-3) + 0.7 (\Tilde{I}_1-3) (\Tilde{I}_2-3) + 0.2 (\Tilde{I}_1-3)^3 + 1.5 (J-1)^2.
            \end{equation}
    \item Gent-Thomas (GT) model \citep{gent_forms_1958}:
        \begin{equation}\label{eq:GT}
            W(\bfF) = 0.5 (\Tilde{I}_1-3) + \log(\Tilde{I}_2/3) + 1.5 (J-1)^2.
            \end{equation}
    \item Arruda-Boyce (AB) model \citep{arruda_three-dimensional_1993}:
            \begin{equation}\label{eq:AB}
                W(\bfF) = 2.5 \sqrt{N_c}\left[ \beta_c \lambda_{c} - \sqrt{N_c}\log \left(\frac{\sinh \beta_c}{\beta_c}\right)\right] - c_\text{AB}  + 1.5 (J-1)^2,
            \end{equation}
             where $\lambda_{c}=\sqrt{{\Tilde{I_1}}/{3}}$, $\beta=\mathcal{L}^{-1}\left({\lambda_{c}}/{\sqrt{N_c}}\right)$, and $\mathcal{L}^{-1}$ denotes the inverse Langevin function. The constants are set to $N_c=28$ (denoting number of polymeric chain segments) and $c_\text{AB}\approx 3.7910$, the latter is used to offset the energy density to zero at $\bfF=\bfI$ (since the Arruda-Boyce feature does not itself vanish for zero deformation).
    \item Ogden model (OG) \citep{ogden_large_1972}:
            \begin{equation}\label{eq:OG}
                W(\bfF) =  \frac{\mu}{\eta}(\lambda_1^{\eta}+\lambda_2^{\eta}+\lambda_3^{\eta}-3) 
            \end{equation}
            where $\lambda_1,\lambda_2,\lambda_3$ are the principal stretches and $\mu=\eta=1.3$.
    \item Anisotropic model with one fiber family at $\alpha_1=45^{\circ}$ orientation (abbreviated as: AI45):
            \begin{equation}\label{eq:AI45}
                W(\bfF) = 0.5(\Tilde{I_1}-3) + 0.75(J-1)^2 + 0.5({\tilde I}_{a_1}-1)^2.
            \end{equation}
    \item Anisotropic model with one fiber family at $\alpha_1=60^{\circ}$ orientation (abbreviated as: AI60):
            \begin{equation}\label{eq:AI60}
               W(\bfF) = 0.5(\Tilde{I_1}-3) + 0.75(J-1)^2 + 0.5({\tilde I}_{a_1}-1)^2.
            \end{equation}
    \item Anisotropic Holzapfel model (HZ) \citep{holzapfel_new_2000} 
            \begin{equation}\label{eq:HZ}
                W(\bfF) = 0.5 (\Tilde{I}_1-3)  + \frac{\kappa_{1}}{2\kappa_{2}}\left[\exp\left(\kappa_{2}({\tilde I}_{a_1}-1)^2\right)+\exp\left(\kappa_{2}({\tilde I}_{a_2}-1)^2\right)-2\right] + (J-1)^2
            \end{equation}
    with two fiber families at $\alpha_1=+30^{\circ}$ and $\alpha_2=-30^{\circ}$ orientations and constants $\kappa_{1} = 0.9$ and $\kappa_{2}=0.8$. 
\end{enumerate}
For benchmarking purposes, the ICNN-based model is then trained to surrogate the aforementioned models by using the NN-EUCLID procedure in Section~\ref{sec:loss}. Further implementation details are presented in \ref{sec:implementation-details}.

To account for the noise in real DIC experiments, we add artificial noise to the synthetic displacement data generated using the FEM simulations. The noise is mainly dependent on the pixel accuracy of the imaging device, which means that the noise level remains constant for every degree of freedom at every load step irrespective of the displacement. The noise is added to the displacement data as
\begin{align}
    u_i^{a,t} = u_i^{\text{fem},a,t} + \eps_i^{a,t} \quad\text{with}\quad \eps_i^{a,t} \sim \mathcal{N}(0,\sigma_u^2) \quad \forall \quad (a,i) \in \mathcal{D},\  t \in \{1,\dots,n_t\},
\end{align}
where $u_i^{\text{fem},a,t}$ and $\eps_i^{a,t}$ denote the true displacement from FEM and  random noise, respectively, the latter sampled independently and identically from a zero-centered normal distribution with standard deviation $\sigma_u$. Following \citet{flaschel_unsupervised_2021}, we consider two noise levels  (normalized relative to specimen length): $\sigma_u=10^{-4}$ (\textit{low noise}) and $\sigma_u=10^{-3}$ (\textit{high noise}) that are representative of modern DIC setups. As per the common practice in DIC, each snapshot of the noisy displacement data is further spatially denoised. For the scope of this work, we use kernel ridge regression (KRR) for denoising; see  \citet{flaschel_unsupervised_2021} for details on the denoising algorithm. The denoised and interpolated displacement data are then used for the ICNN training. 

\subsection{Results}\label{sec:Results}

The detailed ICNN architecture and training protocols for the benchmarks are presented in \ref{sec:implementation-details}. Due to the ill-posedness and sensitivity to initial guess of the NN weights,  we create an ensemble of $n_e=30$ ICNN models with the same architecture but each trained independently with different random weight initialization. The ICNN models with final loss value (see \eqref{loss}) within $20\%$ of the lowest loss value across the ensemble are accepted, while the remaining models are rejected.
The filtering of models with losses higher than the acceptance criterion allows for a qualitative estimate of the confidence associated with the predicted lowest-loss ICNN model. For example, if it is observed that only the lowest-loss model lies within the acceptance criterion, then it can be inferred that the predicted lowest-loss ICNN model has high associated uncertainty. Conversely, if all the ICNN models in the ensemble lie within the acceptance criterion, the predicted lowest-loss ICNN model has high associated certainty. The choice of acceptance criterion also provides a commensurate qualitative estimate of robustness to the NN initialization and does not influence the lowest-loss (best) ICNN model in any way.

\subsubsection{Model accuracy, generalization, and FEM deployment}

\begin{figure}
		\centering
		\text{Benchmark: Strain energy density predictions, low noise ($\sigma_u=10^{-4}$)}\par\medskip
		\includegraphics[width=0.9\textwidth]{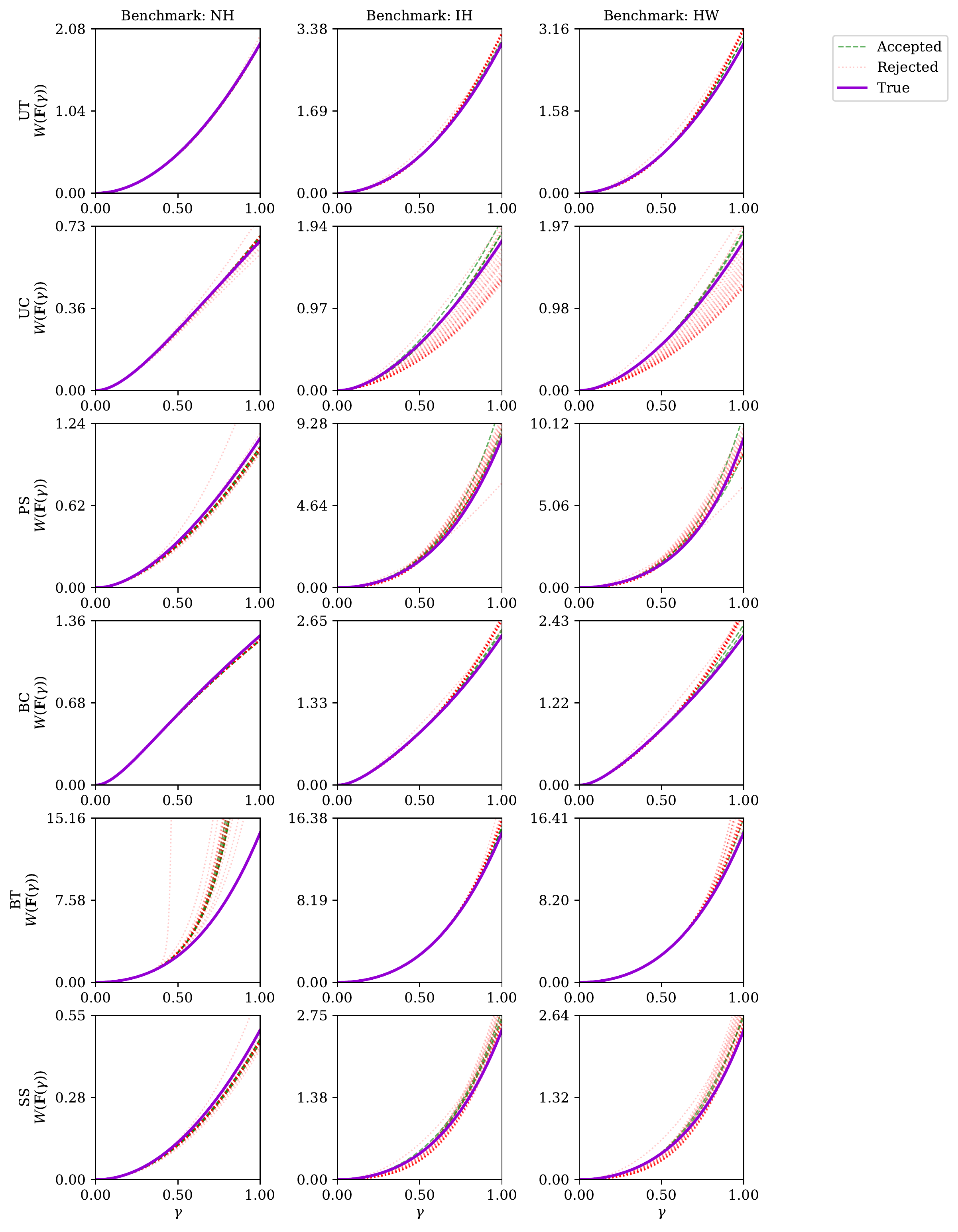}
		\caption{Strain energy density $W(\bfF(\gamma))$ prediction along the different deformation paths in \eqref{eq:strain_paths} using the ICNN-based constitutive models (both accepted and rejected ones) for the case with low noise $(\sigma_u = 10^{-4})$ and benchmarks NH \eqref{eq:NH2}, IH \eqref{eq:IH}, and HW \eqref{eq:HW}. The constitutive response of the (hidden) true model is also shown for reference.}
        \label{fig:NH2_IH_HW_noise=low_W}
\end{figure}

\begin{figure}
		\centering
		\text{Benchmark: Strain energy density predictions, low noise ($\sigma_u=10^{-4}$)}\par\medskip
		\includegraphics[width=0.9\textwidth]{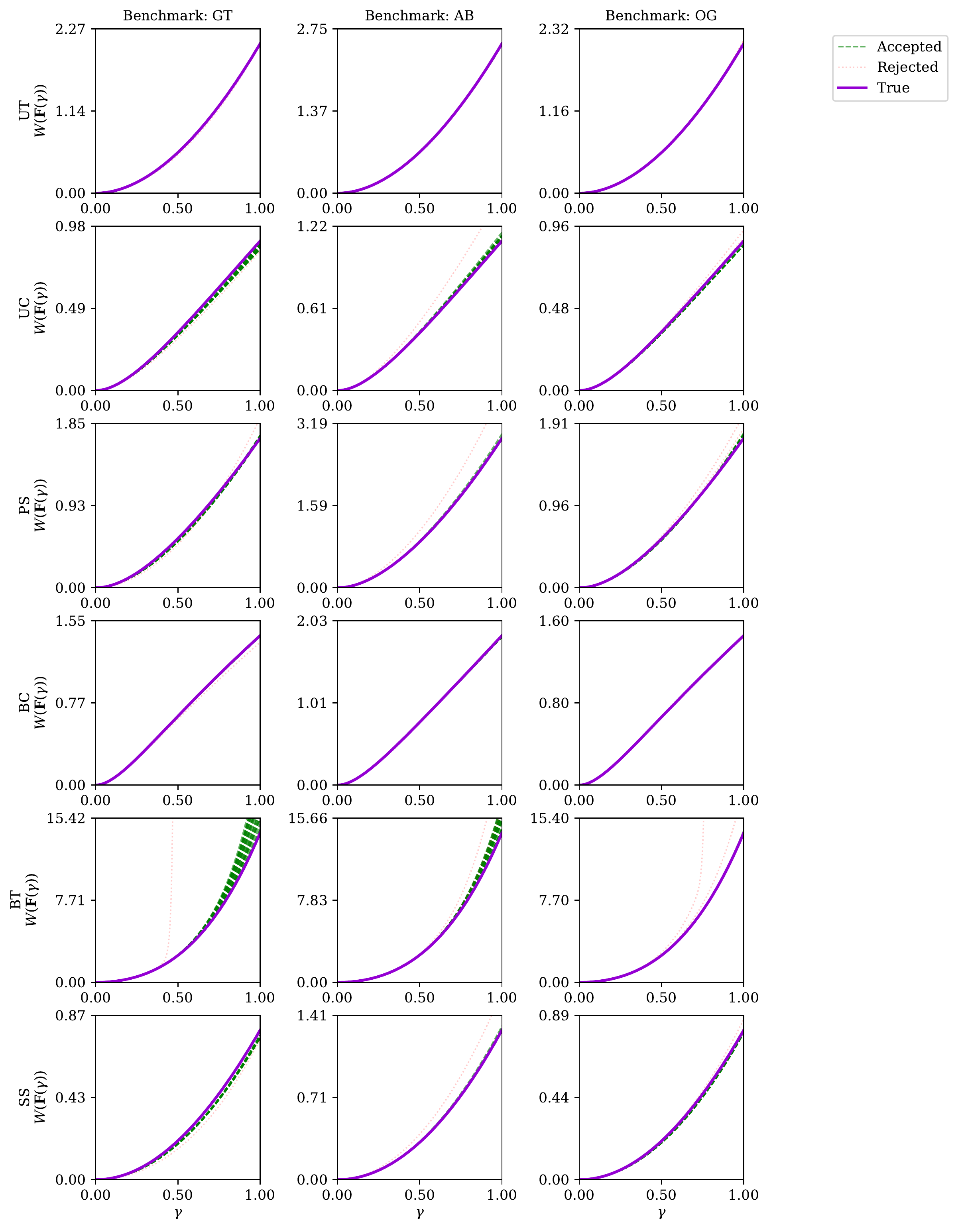}
		\caption{Strain energy density $W(\bfF(\gamma))$ prediction along the different deformation paths in \eqref{eq:strain_paths} using the ICNN-based constitutive models (both accepted and rejected ones) for the case with low noise $(\sigma_u = 10^{-4})$ and benchmarks GT \eqref{eq:GT}, AB \eqref{eq:AB}, and OG \eqref{eq:OG}. The constitutive response of the (hidden) true model is also shown for reference.}\label{fig:GT_AB_OG_noise=low_W}
\end{figure}

\begin{figure}
		\centering
		\text{Benchmark: Strain energy density predictions, high noise ($\sigma_u=10^{-3}$)}
		\includegraphics[width=0.9\textwidth]{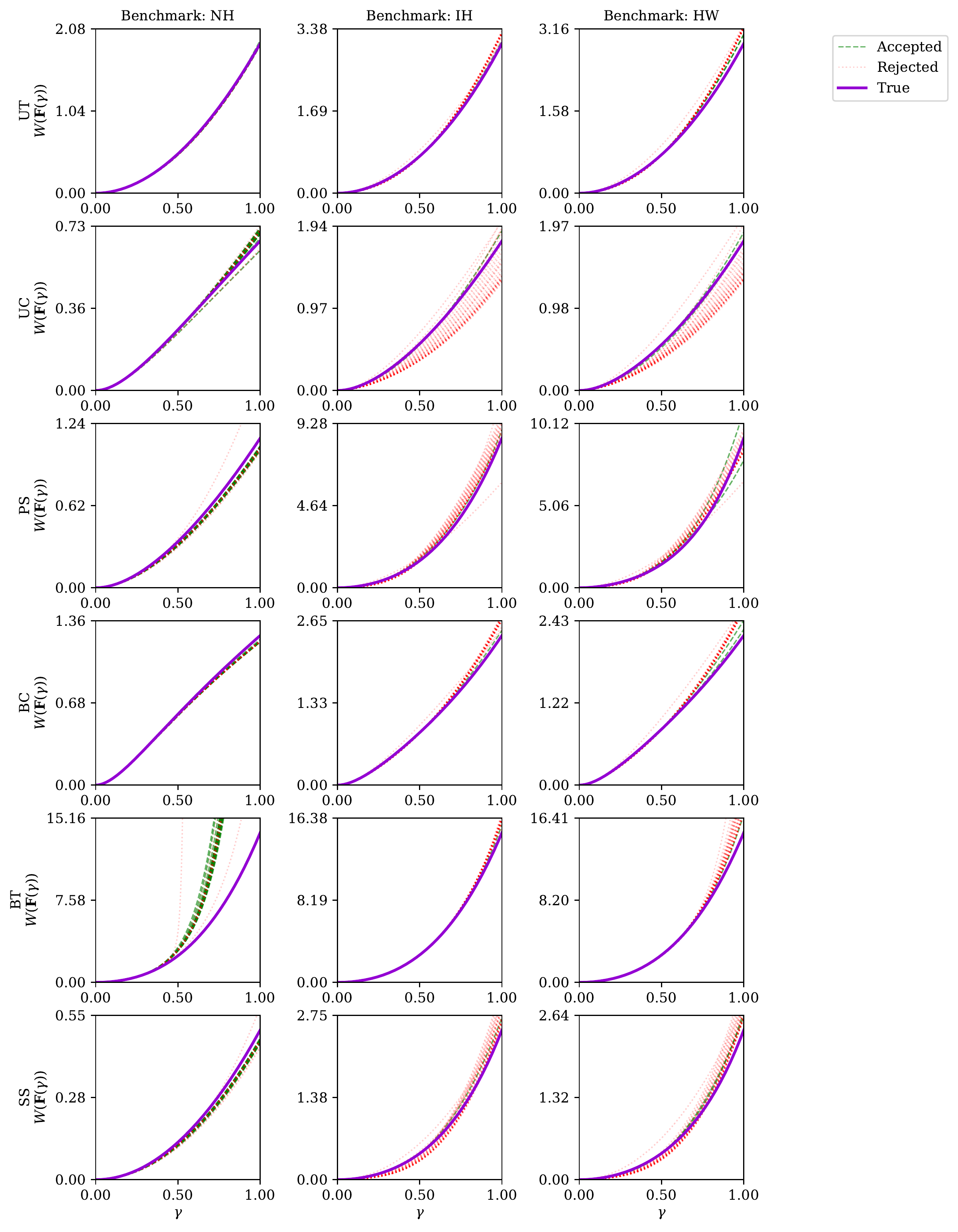}
		\caption{Strain energy density $W(\bfF(\gamma))$ prediction along the different deformation paths in \eqref{eq:strain_paths} using the ICNN-based constitutive models (both accepted and rejected ones) for the case with high noise $(\sigma_u = 10^{-3})$ and benchmarks NH \eqref{eq:NH2}, IH \eqref{eq:IH}, and HW \eqref{eq:HW}. The constitutive response of the (hidden) true model is also shown for reference.}\label{fig:NH2_IH_HW_noise=high_W}
\end{figure}

\begin{figure}
		\centering
		\text{Benchmark: Strain energy density predictions, high noise ($\sigma_u=10^{-3}$)}\par\medskip
		\includegraphics[width=0.9\textwidth]{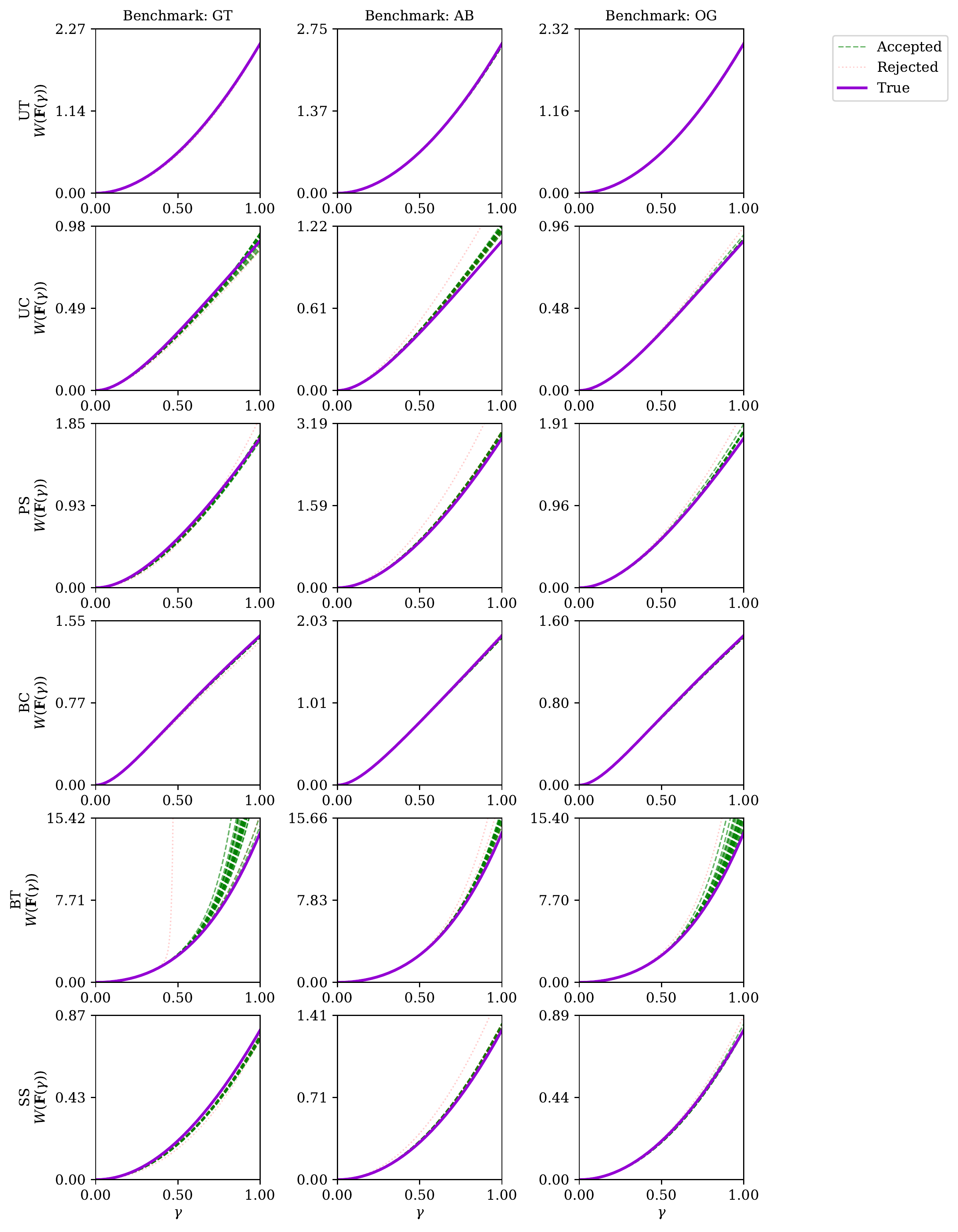}
		\caption{Strain energy density $W(\bfF(\gamma))$ prediction along the different deformation paths in \eqref{eq:strain_paths} using the ICNN-based constitutive models (both accepted and rejected ones) for the case with high noise $(\sigma_u = 10^{-3})$ and benchmarks GT \eqref{eq:GT}, AB \eqref{eq:AB}, and OG \eqref{eq:OG}. The constitutive response of the (hidden) true model is also shown for reference.}\label{fig:GT_AB_OG_noise=high_W}
\end{figure}

For all the benchmarks \eqref{eq:NH2}-\eqref{eq:HZ}, we evaluate the  ICNN-based constitutive models against the ground truth models  along six deformation paths: uniaxial tension (UT), uniaxial compression (UC), biaxial tension (BT), biaxial compression (BC), simple shear (SS) and pure shear (PS). The deformation gradients for these paths are given by:
\be
\begin{aligned}
\boldsymbol{F}^{\text{UT}}(\gamma) = \begin{bmatrix}
1 + \gamma & 0\\
0 & 1
\end{bmatrix}, \ \ \
\boldsymbol{F}^{\text{UC}}(\gamma) = \begin{bmatrix}
\frac{1}{1 + \gamma} & 0\\
0 & 1
\end{bmatrix}, \ \ \ 
\boldsymbol{F}^{\text{BT}}(\gamma) = \begin{bmatrix}
1 + \gamma & 0\\
0 & 1+ \gamma
\end{bmatrix}, \\
\boldsymbol{F}^{\text{BC}}(\gamma) = \begin{bmatrix}
\frac{1}{1 + \gamma} & 0 \\
0 & \frac{1}{1 + \gamma}
\end{bmatrix}, \ \ \
\boldsymbol{F}^{\text{SS}}(\gamma) = \begin{bmatrix}
1 & \gamma\\
0 & 1
\end{bmatrix}, \ \ \
\boldsymbol{F}^{\text{PS}}(\gamma) = \begin{bmatrix}
1 + \gamma & 0\\
0 & \frac{1}{1 + \gamma}
\end{bmatrix},
\end{aligned}
\label{eq:strain_paths}
\ee
where $\gamma\in[0,1]$ is a loading parameter. Note that these deformation paths are purely for evaluation purposes and in no way contribute to model training. In all the benchmarks, the strain energy density predictions from the accepted ICNN models -- despite the multiplicity in solutions due to the ill-posedness -- show good agreement with the ground truth; see Figures \ref{fig:NH2_IH_HW_noise=low_W}, \ref{fig:GT_AB_OG_noise=low_W}, \ref{fig:AI45_AI60_HA_noise=low_W} and Figures \ref{fig:NH2_IH_HW_noise=high_W}, \ref{fig:GT_AB_OG_noise=high_W}, \ref{fig:AI45_AI60_HA_noise=high_W} for the low and high noise cases, respectively. The ICNN models also show good agreement with the ground truth in terms of the first Piola-Kirchhoff stress along the six deformation paths for both noise levels; see Figures \ref{fig:NH2_IH_HW_noise=low_Pij}-\ref{fig:AI45_AI60_HA_noise=high_Pij}. 

\begin{figure}[t]
		\centering
		\includegraphics[width=1.0\textwidth]{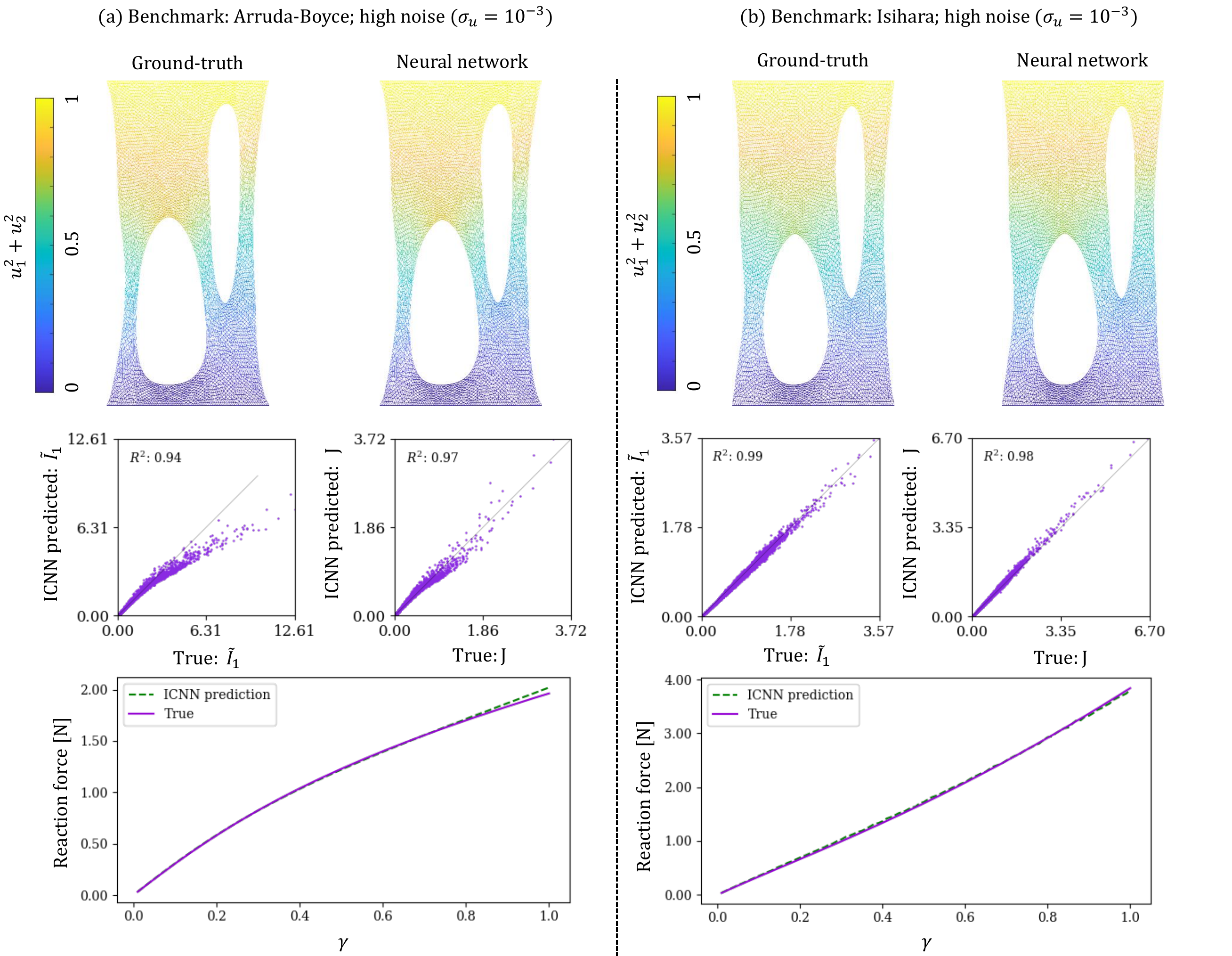}
		\caption{Comparison of FEM solutions for the validation specimen (\figurename\ref{fig:geometries}b) between the ground-truth and ICNN-based constitutive models for (a) Arruda-Boyce \eqref{eq:AB} and (b) Isihara \eqref{eq:IH} benchmarks. 
		(\textit{Top}) Deformed shape (shaded by displacement magnitude) of the specimen at loading parameter $\delta=1$ (see \figurename\ref{fig:geometries}b for details). 
		(\textit{Middle}) Predicted strain invariants ($\Tilde{I}_1$ and $J$) at each quadrature point from the ICNN-based simulation vs. true strain invariants (from ground-truth simulation) in the validation specimen. In the ideal case, we expect each data point to lie on a line (indicated as red dashed line) with zero intercept and unit slope for perfect accuracy. The coefficient of determination ($R^2$) with respect to this line serves as a measure of accuracy.
		(\textit{Bottom}) Comparison of the vertical reaction force on the top surface as a function of loading parameter $\delta$, obtained from the ICNN-based simulation vs.~the ground-truth simulation.}\label{fig:fem-validation}
\end{figure}

For further validation, we deploy the ICNN-based constitutive model  within a finite element simulation framework. Without loss of generality, we use linear triangular elements and a Newton-Raphson-based nonlinear solver, which requires tangent computations. The element-wise  stress and tangent modulus are computed using automatic differentiation via  \eqref{eq:PK_ansatz} and \eqref{eq:tangent}, respectively. To ensure that the ICNN models generalize to a simulation that is different from the one they are trained on, we consider a validation specimen (\figurename \ref{fig:geometries}b) with a more complex geometry than the training specimen,  featuring two asymmetric elliptical holes  and subjected to quasi-static uniaxial loading. For these validation simulations we use the ICNN material models with the lowest loss across the trained ensembles.  \figurename\ref{fig:fem-validation} shows that the solutions to the above mechanical boundary value problem for two representative ground-truth constitutive models (Arruda-Boyce \eqref{eq:AB} and Isihara \eqref{eq:IH}) are in excellent agreement with those based on their respective ICNN ensemble-based constitutive models. This is quantitatively supported by good accuracy in the element-wise strain invariants (coefficient of determination or $R^2$ scores greater than 0.94 in the high noise case) as well as in the reaction forces between ground-truth and ICNN-based simulations. The ICNN architecture and smooth activations also mitigate spurious, oscillatory, and non-smooth artifacts in the overall stress-strain response and ensure smooth first and second derivatives of the strain energy density for robust deployment in FEM setting. In \ref{sec:non-convex-results}, we demonstrate that a simple feed-forward NN without the input-convex architecture and smoothness is not suitable for constitutive modeling.

\begin{figure}[t]
		\centering
		\includegraphics[width=1.0\textwidth]{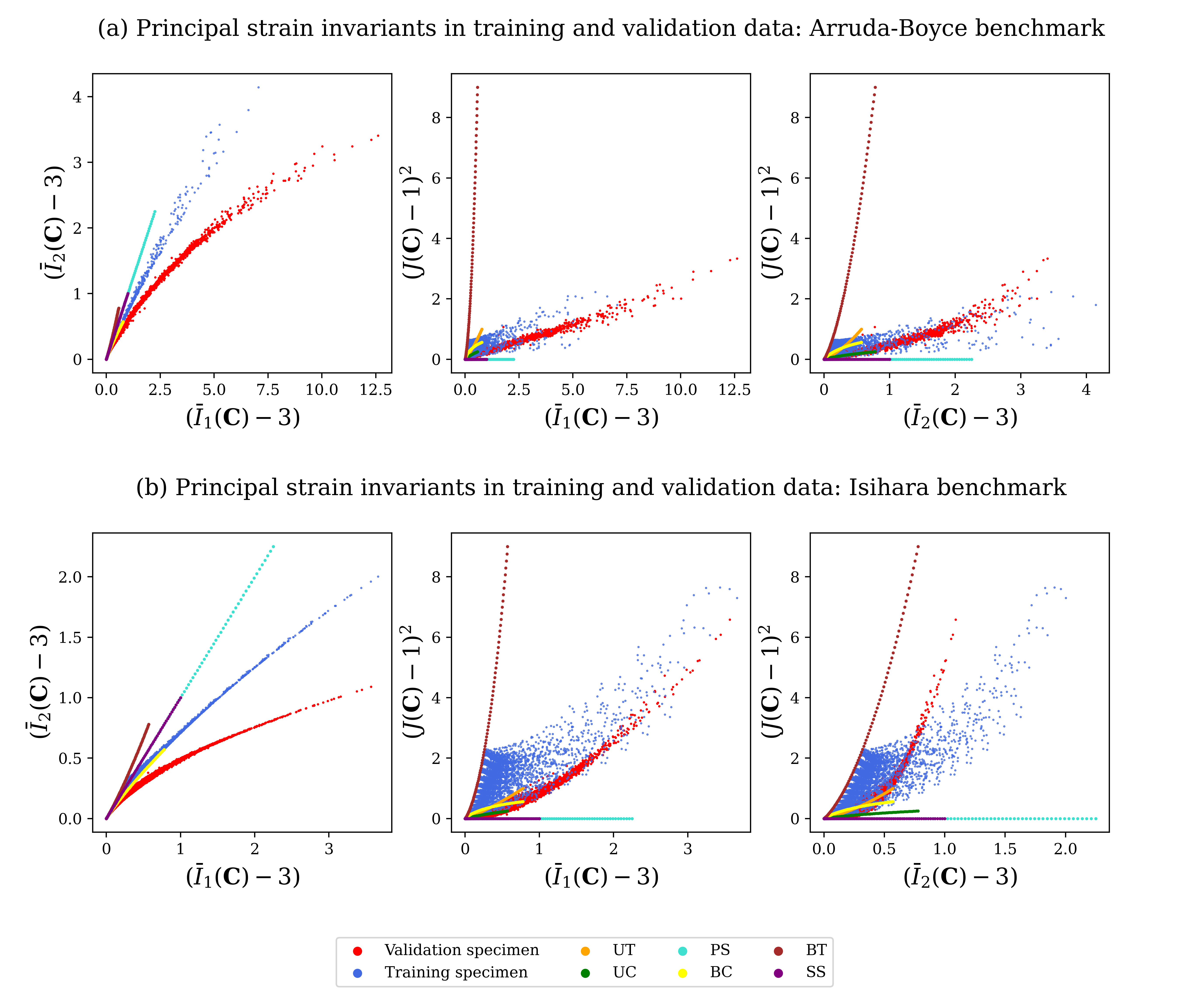}
		\caption{Two-dimensional projections of the principal strain invariants $(\Tilde{I}_1-3)$, $(\Tilde{I}_2-3)$, and $(J-1)^2$ at all the elements in the training specimen (\figurename\ref{fig:geometries}a) across all the loadsteps for (a) Arruda-Boyce \eqref{eq:AB} and (b) Isihara \eqref{eq:IH} benchmarks. Also shown are the principal strain invariants of the validation specimen (\figurename\ref{fig:geometries}b) across all elements and loadsteps and the six evaluation deformation paths \eqref{eq:strain_paths}. Note that all the data shown here are based on ground-truth model without noise (including strain invariants of the validation specimen.)}\label{fig:invariants-space_AB_IH}
\end{figure}

For visualizing the generalization capability of the ICNN models, we plot the two-dimensional projections of the principal strain invariants in the training specimen (across all load steps and all elements) -- as demonstrated in \figurename\ref{fig:invariants-space_AB_IH}. The strain invariants generated by the six deformation paths in \eqref{eq:strain_paths} as well as from the validation specimens (across all load steps and all elements) are also added to the same plot. This confirms the generalization capability of the ICNN-based constitutive models, as the validation strain states go beyond the space of the strain states observed in the training specimen. Uniaxial tension, biaxial compression and uniaxial compression deformation paths are mostly encompassed by the training data, yielding almost perfect results for all the benchmarks. It also explains the significant mismatch between the ground truth and predicted strain energy densities -- particularly for $\gamma>0.5$ in simple shear, pure shear, and biaxial tension deformations -- as these deformations are too large relative to those observed in the training specimen.

\subsubsection{Learning hidden anisotropy}
\begin{figure}
		\centering
		\text{Benchmark: Strain energy density predictions, low noise ($\sigma_u=10^{-4}$)}\par\medskip
		\includegraphics[width=0.9\textwidth]{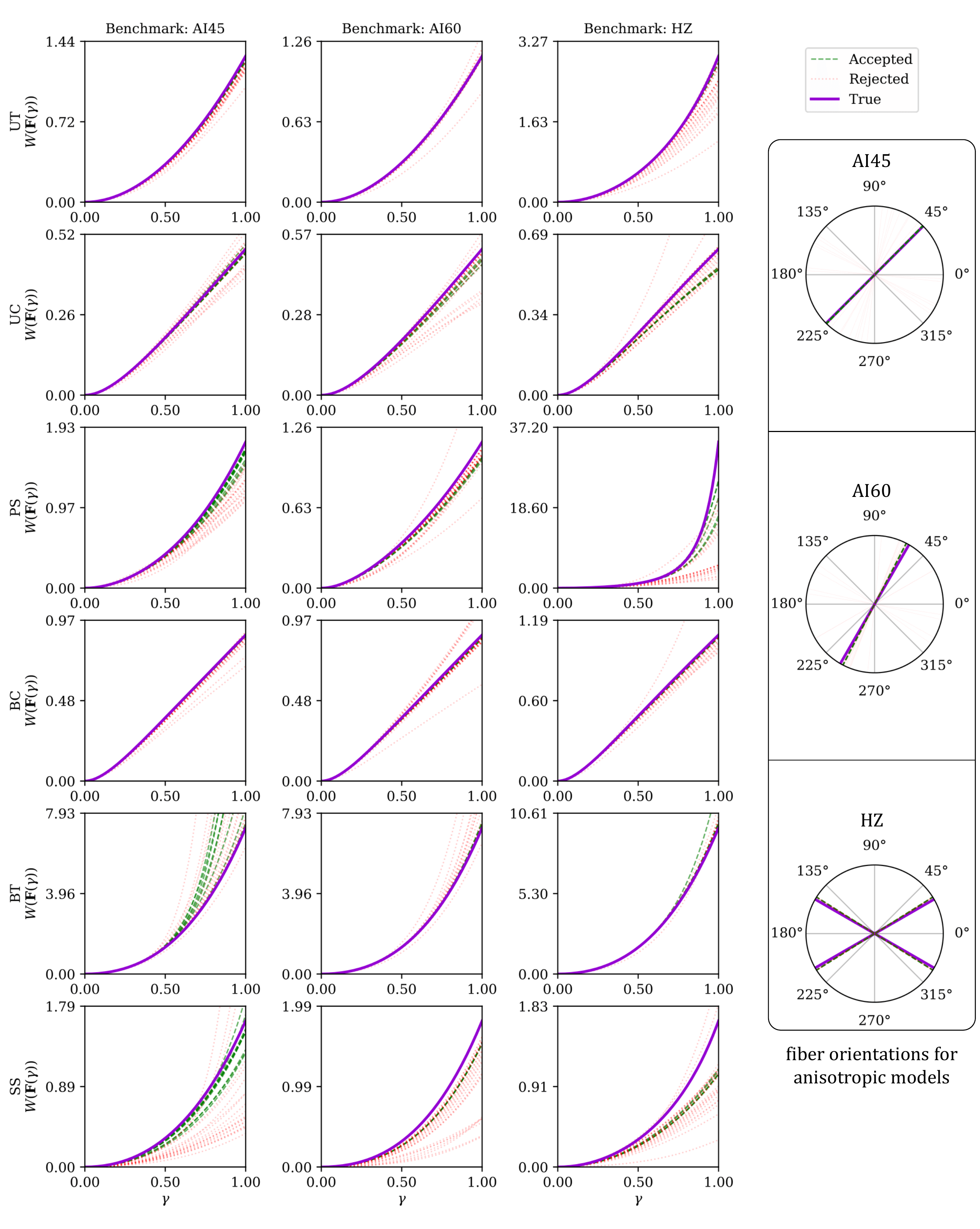}
		\caption{Strain energy density $W(\bfF(\gamma))$ prediction along the different deformation paths in \eqref{eq:strain_paths} using the ICNN-based constitutive models (both accepted and rejected ones) for the case with low noise $(\sigma_u = 10^{-4})$ and benchmarks AI45 \eqref{eq:AI45}, AI60 \eqref{eq:AI60}, and HZ \eqref{eq:HZ}. The constitutive response of the (hidden) true model is also shown for reference. The insets show the fiber orientations discovered by the ICNN-based models and the fiber orientations in the (hidden) true model.}\label{fig:AI45_AI60_HA_noise=low_W}
\end{figure}

\begin{figure}
		\centering
		\text{Benchmark: Strain energy density predictions, high noise ($\sigma_u=10^{-3}$)}\par\medskip
		\includegraphics[width=0.9\textwidth]{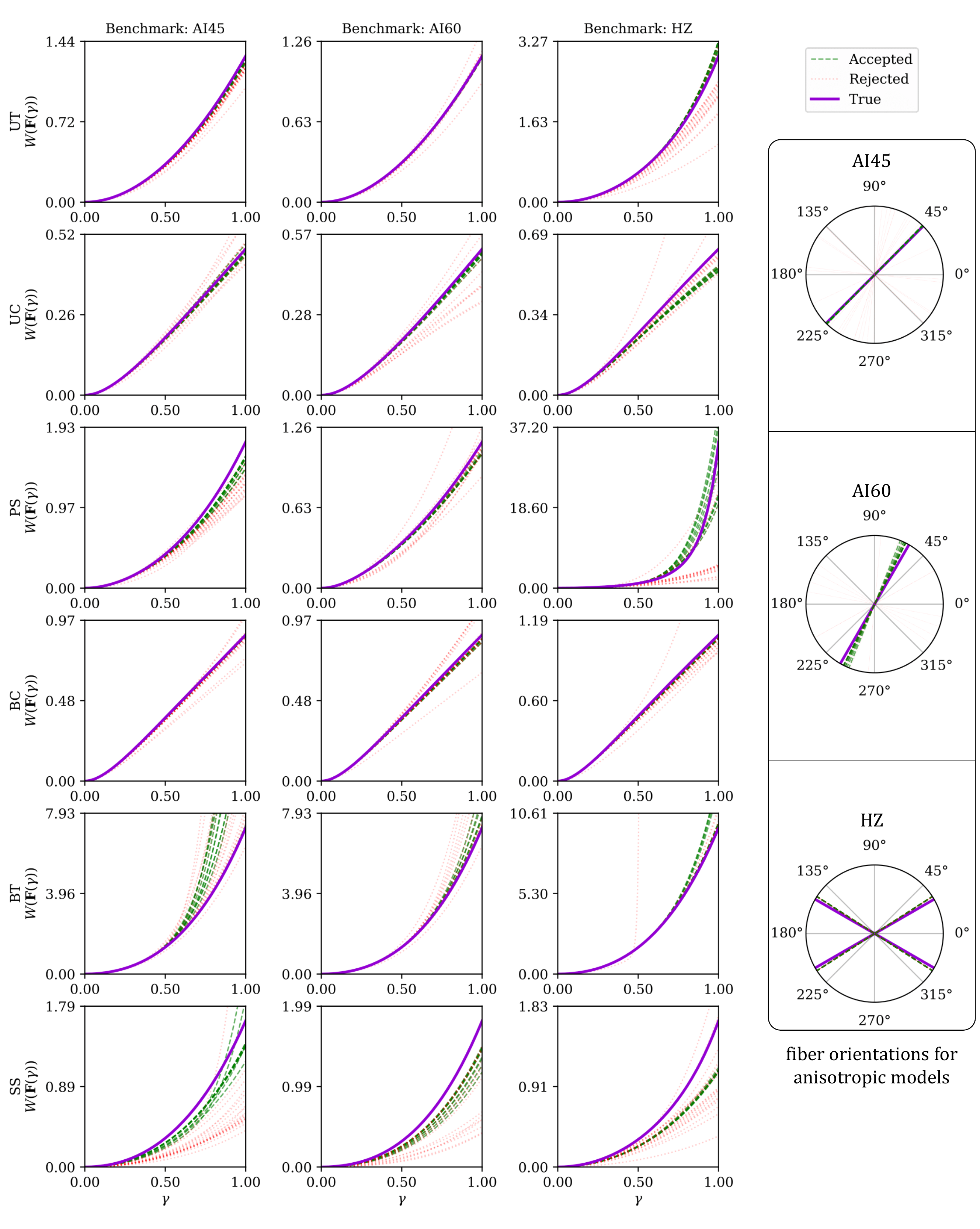}
		\caption{Strain energy density $W(\bfF(\gamma))$ prediction along the different deformation paths in \eqref{eq:strain_paths} using the ICNN-based constitutive models (both accepted and rejected ones) for the case with high noise $(\sigma_u = 10^{-3})$ and benchmarks AI45 \eqref{eq:AI45}, AI60 \eqref{eq:AI60}, and HZ \eqref{eq:HZ}. The constitutive response of the (hidden) true model is also shown for reference. The insets show the fiber orientations discovered by the ICNN-based models and the fiber orientations in the (hidden) true model.}\label{fig:AI45_AI60_HA_noise=high_W}
\end{figure}

For the benchmarks with single fiber family (\eqref{eq:AI45} and \eqref{eq:AI60}) and Holzapfel model \eqref{eq:HZ} with two fiber families, the fiber orientations in $\calA$ are not known a priori but rather learned by the ICNNs as part of the training (see Algorithm~\ref{alg:pseudocode} for pseudocode). Despite the unsupervised setting,  the fiber orientations in  $\calA$ are accurately identified in both low and high noise cases -- as demonstrated in Figures \ref{fig:AI45_AI60_HA_noise=low_W} and \ref{fig:AI45_AI60_HA_noise=high_W}, respectively.

\section{Conclusion and Outlook}\label{sec:Conclusion}
We developed NN-EUCLID, i.e. an unsupervised learning approach for the encoding of material models in deep NNs with no reliance on stress data, and demonstrated its performance for isotropic and anisotropic hyperelasticity. The proposed framework only requires realistically measurable data, i.e., full-field displacements and global forces. The constitutive model is based on ICNNs whose special structure guarantees the physical requirements of material objectivity, (local) material stability, and stress-free reference configuration. The lack of stress labels in the unsupervised learning setting is  compensated for by minimizing a physics-motivated loss which encapsulates the constraint that the observed displacement fields must satisfy the conservation of linear momentum. Through several benchmarks for isotropic and anisotropic hyperelasticity, we demonstrated the capability of accurately learning the underlying material behavior by only providing data from a single experiment under different noise levels that are representative of contemporary DIC setups. The ICNN framework also automatically discovers unknown  principal directions of anisotropy (fiber orientations) in anisotropic hyperelastic materials. We further showed that the ICNN-based constitutive models can generalize to strain states and specimen geometries beyond those of the training specimen; and that they can be deployed within FEM simulations involving both stress and tangent computations with good accuracy (relative to the otherwise unknown ground-truth constitutive model).

In a broader view, the proposed framework offers a departure from the reliance on stress labels for deep learning of constitutive models. Future developments include experimental validation as well as the extension to unsupervised deep learning of inelastic material behavior.




\section*{Acknowledgments}

MF and LDL would like to acknowledge funding by SNF through grant N. $200021\_204316$ ``Unsupervised data-driven discovery of material laws''.

\section*{Declaration of competing interest}
The authors declare that they have no known competing financial interests or personal relationships that could have appeared to influence the work reported in this paper.

\section*{Code availability}
The codes developed in current study will be freely open-sourced at the time of publication.

\section*{Data availability}
The data generated in current study will be made freely available at the time of publication.

\bibliographystyle{elsarticle-harv}
\bibliography{Bib}

\clearpage

\appendix

\section{Implementation details}\label{sec:implementation-details}

All the parameters and hyperparameters used in data generation and NN training are listed in Table~\ref{tab:parameters} with the following related explanations.

\begin{table}[]
\centering
\caption{List of parameters and hyperparameters used for the data generation and benchmarks.} 
\label{tab:parameters}
\begin{tabular}{lcc}
\hline
\multicolumn{1}{l}{\textbf{Parameter}} & \textbf{Notation} & \textbf{Value} \\ \hline
\textit{Training specimen:}\\
$\quad$ Number of nodes in mesh for FEM-based data generation  & - & $63,601$  \\
$\quad$ Number of nodes in data available for learning  & $n_n$ & $1,441$\\
$\quad$ Number of reaction force constraints & $n_{\beta}$   & $4$   \\ 
$\quad$ Number of data snapshots for NH, GT    & $n_{t}$     & 3 \\
$\quad$ Number of data snapshots for IH, HW, AI45, AI60, HZ    & $n_{t}$     & 8 \\
$\quad$ Number of data snapshots for AB    & $n_{t}$     & 10 \\
$\quad$ Number of data snapshots for OG    & $n_{t}$     & 6 \\

$\quad$ Loading parameter for NH, GT, IH, HW  & $\delta$  & $\{0.1 \times t: t=1,\dots,n_t\}$ \\
$\quad$ Loading parameter for AB, OG, AI45, AI60, HZ  & $\delta$  & $\{0.05 \times t: t=1,\dots,n_t\}$ \\ \hline
\textit{Validation specimen:}\\
$\quad$ Number of nodes in the mesh  & - & $4,908$ \\
$\quad$ Loading parameter   & $\delta$  & $\{0.01 \times t: t=1,\dots,100\}$ \\ \hline
\textit{ICNN hyperparameters:}\\
$\quad$ Number of hidden layers & $N-1$    & $3$\\
$\quad$ Number of neurons in each hidden layer $(k=1,\dots,N-1)$  & $d_k$    & $64$  \\
$\quad$ Dropout rate in hidden layers  & - & $20\%$      \\
$\quad$ Scaling parameter in $\calG$ &$c_\calG$&    $1.0$      \\
$\quad$ Scaling parameter in $\calF$ &$c_\calF$&    $1/12$      \\
$\quad$ Number of models in the ensemble &$n_e$&    $30$      \\
$\quad$ Number of epochs & - &    $500$      \\
$\quad$ Learning rate schedule &-&    cyclic      \\
$\quad$ Base learning rate & - &    $0.001$      \\
$\quad$ Maximum learning rate & - &    $0.1$      \\
$\quad$ Learning rate cycle upward steps & - &    $50$      \\
$\quad$ Learning rate cycle downward steps & - &    $50$      \\
\end{tabular}
\end{table}

\subsection{Data generation}
All lengths and displacements are normalized with respect to the side length of the undeformed specimen. The training specimen is discretized with a high-resolution mesh (same as \citet{flaschel_unsupervised_2021}) consisting of 63,601 nodes. The noisy displacements of the training specimen are spatially-denoised using the same KRR denoiser and parameters as in \citet{flaschel_unsupervised_2021}. For data efficiency, the denoised displacements are then projected onto a coarser mesh with $n_n = 1,441$ nodes, which are used for training the ICNN models.

\subsection{ICNN}
The  ICNN architecture consists of three hidden layers ($N=4$) with 64 neurons each ($d_1=d_2=d_3=64$). We use the Adam optimizer \citep{Kingma2014}  with automatic differentiation-based backpropagation to train the ICNN parameters $\calQ$ and $\calA$. The training is carried out for 500 epochs, which was observed to be sufficient for the loss \eqref{loss} to converge an acceptable value. To mitigate the network parameters getting stuck in bad local minima, the learning rate of the optimizer is linearly cycled from $0.001$ to $0.1$ and back every 100 epochs. Dropout rate of $20\%$ is used as regularization to prevent overfitting during training \citep{Srivastava2014}. Note that dropout is turned off for evaluation during validation/testing. The scaling coefficients in \eqref{activations} are set to $c_\calG=1.0$ and $c_\calF=1/12$ to avoid vanishing and/or exploding gradients. 

\subsection{Anisotropy}
For anisotropic hyperelasticity, the NN-based constitutive model requires  additional angle parameters $\calA=\{\alpha_1,\alpha_2,\dots\}$ to compute the anisotropic invariants $\{{\tilde I}^a_1,{\tilde I}^a_2,\dots\}$. Since $\alpha_i$ is constrained to $[0,\pi)$, this calls for constrained optimization, which is challenging, particularly with NNs. To avoid this, we first initialize a trainable parameter $\zeta_i \in \mathbb{R}$. The fiber orientation is then given by
\be
\alpha_i=\frac{\pi}{1+e^{-\zeta_i}}.
\ee
This change of variables from $\alpha_i$ to $\zeta_i$ allows unconstrained optimization, which is easier than constrained optimization.

\section{Pseudocode for unsupervised ICNN training}

Algorithm \ref{alg:pseudocode} summarizes the unsupervised training of the ICNN-based constitutive models.

\begin{algorithm} 
\caption{Unsupervised training of the ICNN-based constitutive models}
\label{alg:pseudocode}
\begin{algorithmic}[1]
\State \textbf{Input:} Point-wise displacement data $\calU = \{\bfu^{a,t} \in \mathbb{R}^2 : a = 1, \dots, n_n; t=1,\dots,n_t\}$ 
\State \textbf{Input:} Global reaction forces $\{R^{\beta,t}: \beta=1,\dots,n_\beta; t=1,\dots,n_t\}$
\State \text{Randomly initialize ICNN parameters: $\calQ,\calA$.}
\State \text{Initialize learning rate scheduler}
\State \text{Initialize Adam optimizer with parameters $\calQ,\calA$ and learning rate scheduler}
    \For{$e = 1,\dots,n_e$} \Comment{Training epochs}
        \State Loss: $\ell\gets 0$ \Comment{Initialize loss for current epoch}
        \For{$t = 1,\dots,n_t$} \Comment{Iterate over snapshots}
            \For{each element in mesh}
                \State $W^{0} \gets W^{\text{NN}}_{\calQ,\calA}\vert_{\bfF=\bfI}$ \Comment{see \eqref{zero-energy}}
                \State $\bfH \gets -\left.\frac{\partial W^{\text{NN}}_{\calQ,\calA}}{\partial \bfF}\right\vert_{\bfF=\bfI}$ \Comment{see \eqref{H-matrix}}
                \State $W \gets W^{\text{NN}}_{\calQ,\calA} +  W^{0} + \bfH:\bfE $\Comment{see \eqref{eq:ansatz}}
                \State  $\boldsymbol{P} \gets \frac{\partial  W}{\partial \boldsymbol{F}}$\Comment{see \eqref{eq:PK_ansatz}}
                
            \EndFor
            \For{$a = 1,\dots,n_a$}
                \For{$i = 1,2$}
	                \State Compute force $f^{a,t}_i$ using \eqref{reducedWeakForm}
                \EndFor
            \EndFor
            \For{$(a,i) \in  \calD^\text{free}$}
                \State $\ell\gets\ell + \left(f^{a,t}_i\right)^2 $ \Comment{force balance at free degrees of freedom; see \eqref{loss}}
            \EndFor
            \For{$\beta= 1,\dots, n_\beta$}
                \State $r^{\beta,t} \gets 0$
                \For{$(a,i) \in  \calD^\text{fix}_\beta$}
                    \State $r^{\beta,t} \gets r^{\beta,t} + f^{a,t}_i$ \Comment{see \eqref{loss}}
                \EndFor
                \State $\ell\gets \ell + \left(R^{\beta,t}-r^{\beta,t}\right)^2$ \Comment{force balance at fixed degrees of freedom; see \eqref{loss}}
            \EndFor
        \EndFor
        \State Compute gradients $\partial \ell/\partial \calQ$ and $\partial \ell/\partial \calA$ using automatic differentiation
        \State Update $\calQ$ and $\calA$ with Adam optimizer using gradients $\partial \ell/\partial \calQ$ and $\partial \ell/\partial \calA$
        \State Update learning rate with learning rate scheduler based on epoch number $e$ 
    \EndFor
    \State \textbf{Output:} Trained ICNN model $W^\text{NN}_{\calQ,\calA}$
\end{algorithmic}
\end{algorithm}

\section{Model accuracy in stress predictions}

\begin{figure}
		\centering
		\text{Benchmark: First Piola-Kirchhoff stress component predictions, low noise ($\sigma_u=10^{-4}$)}\par\medskip
		\includegraphics[width=0.9\textwidth]{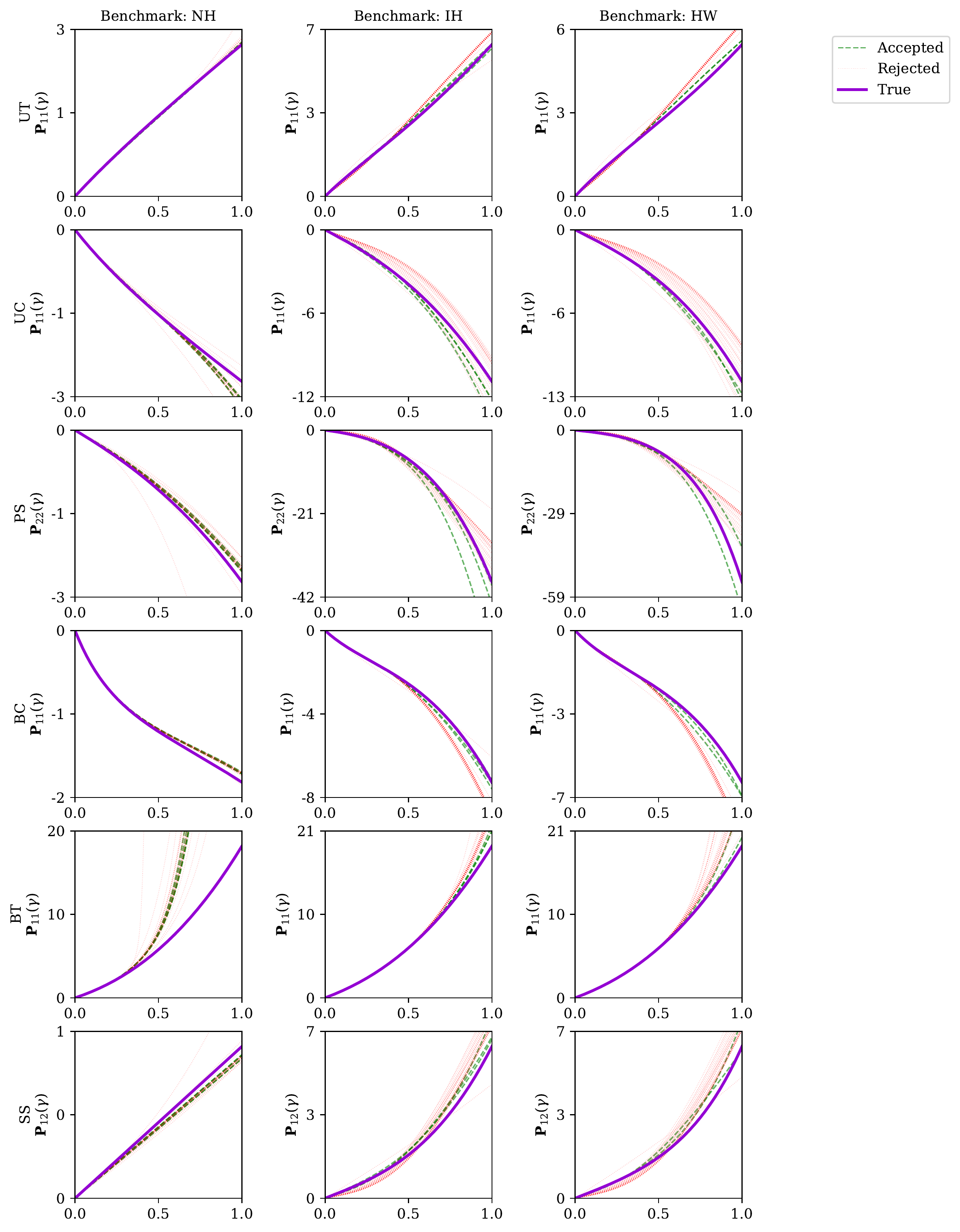}
		\caption{First Piola-Kirchhoff stress $\bfP(\bfF(\gamma))$ component prediction along the different deformation paths in \eqref{eq:strain_paths} using the ICNN-based constitutive models (both accepted and rejected ones) for the case with low noise $(\sigma_u = 10^{-4})$ and benchmarks NH \eqref{eq:NH2}, IH \eqref{eq:IH}, and HW \eqref{eq:HW}. The constitutive response of the (hidden) true model is also shown for reference.}
		\label{fig:NH2_IH_HW_noise=low_Pij}
\end{figure}

\begin{figure}
		\centering
		\text{Benchmark: First Piola-Kirchhoff stress component predictions, low noise ($\sigma_u=10^{-4}$)}\par\medskip
		\includegraphics[width=0.9\textwidth]{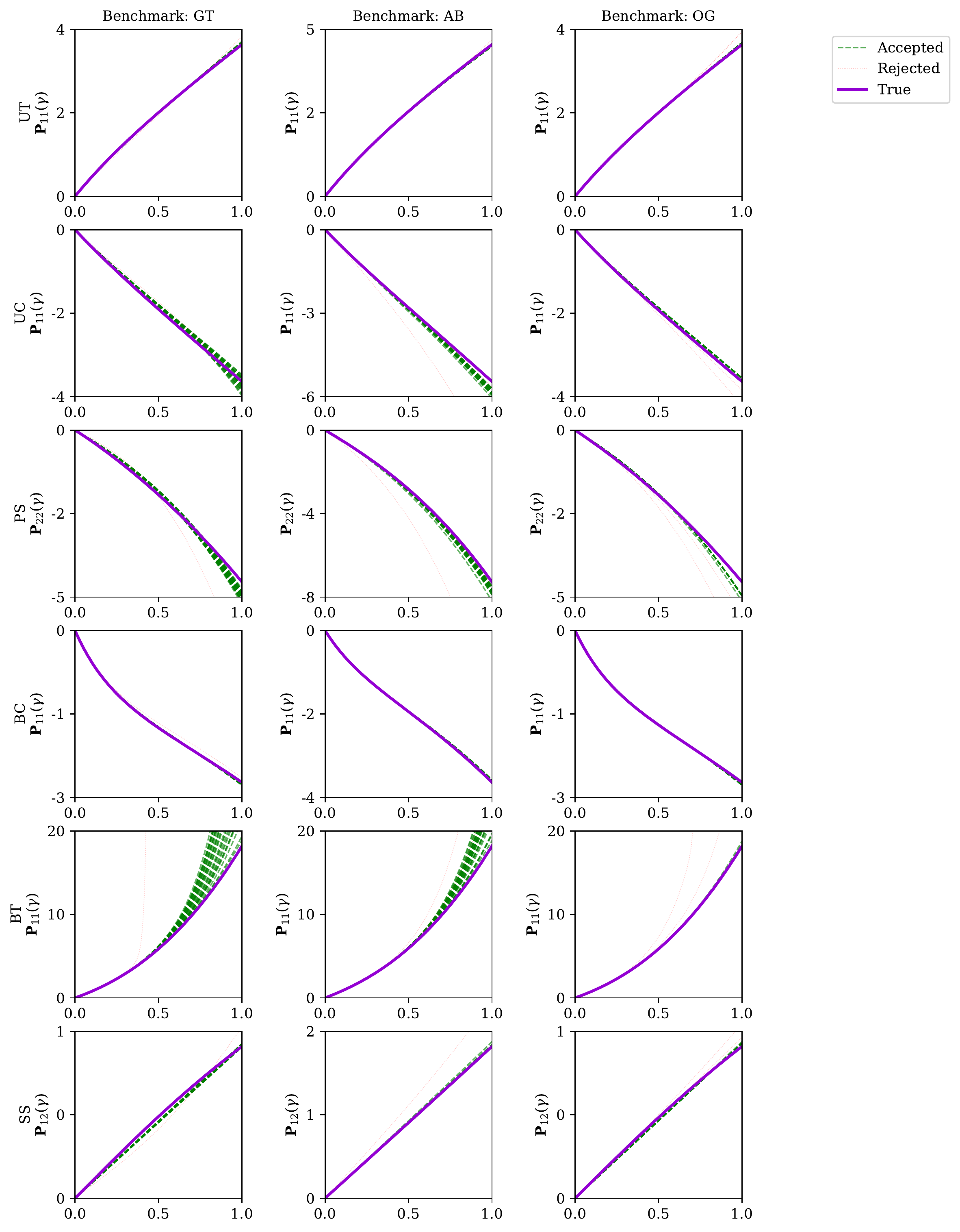}
		\caption{First Piola-Kirchhoff stress $\bfP(\bfF(\gamma))$ component prediction along the different deformation paths in \eqref{eq:strain_paths} using the ICNN-based constitutive models (both accepted and rejected ones) for the case with low noise $(\sigma_u = 10^{-4})$ and benchmarks GT \eqref{eq:GT}, AB \eqref{eq:AB}, and OG \eqref{eq:OG}. The constitutive response of the (hidden) true model is also shown for reference.}\label{fig:GT_AB_OG_noise=low_Pij}
\end{figure}

\begin{figure}
		\centering
		\text{Benchmark: First Piola-Kirchhoff stress component predictions, low noise ($\sigma_u=10^{-4}$)}\par\medskip
		\includegraphics[width=0.9\textwidth]{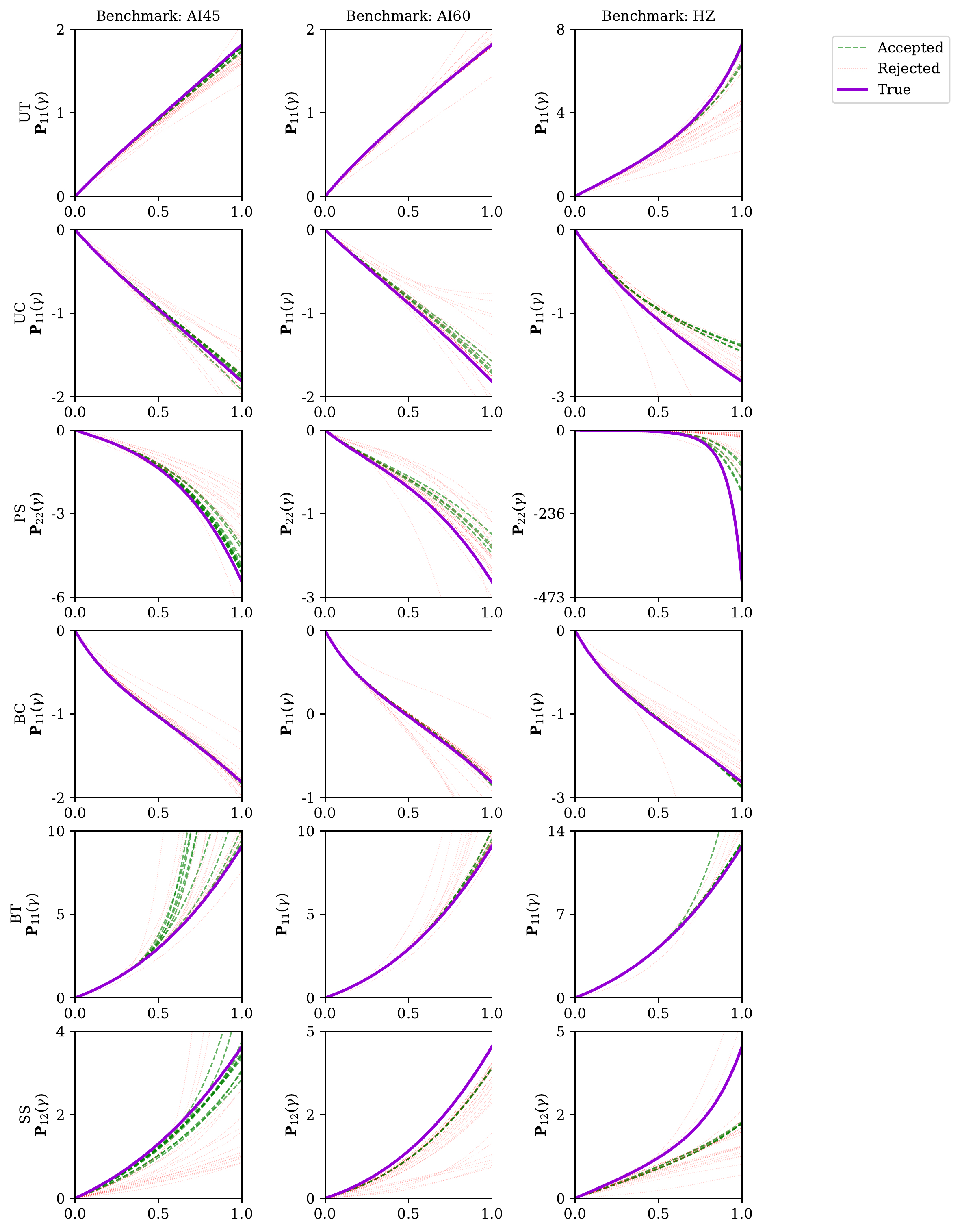}
		\caption{First Piola-Kirchhoff stress $\bfP(\bfF(\gamma))$ component prediction along the different deformation paths in \eqref{eq:strain_paths} using the ICNN-based constitutive models (both accepted and rejected ones) for the case with low noise $(\sigma_u = 10^{-4})$ and benchmarks AI45 \eqref{eq:AI45}, AI60 \eqref{eq:AI60}, and HZ \eqref{eq:HZ}. The constitutive response of the (hidden) true model is also shown for reference.}\label{fig:AI45_AI60_HA_noise=low_Pij}
\end{figure}

\begin{figure}
		\centering
		\text{Benchmark: First Piola-Kirchhoff stress component predictions, high noise ($\sigma_u=10^{-3}$)}\par\medskip
		\includegraphics[width=0.9\textwidth]{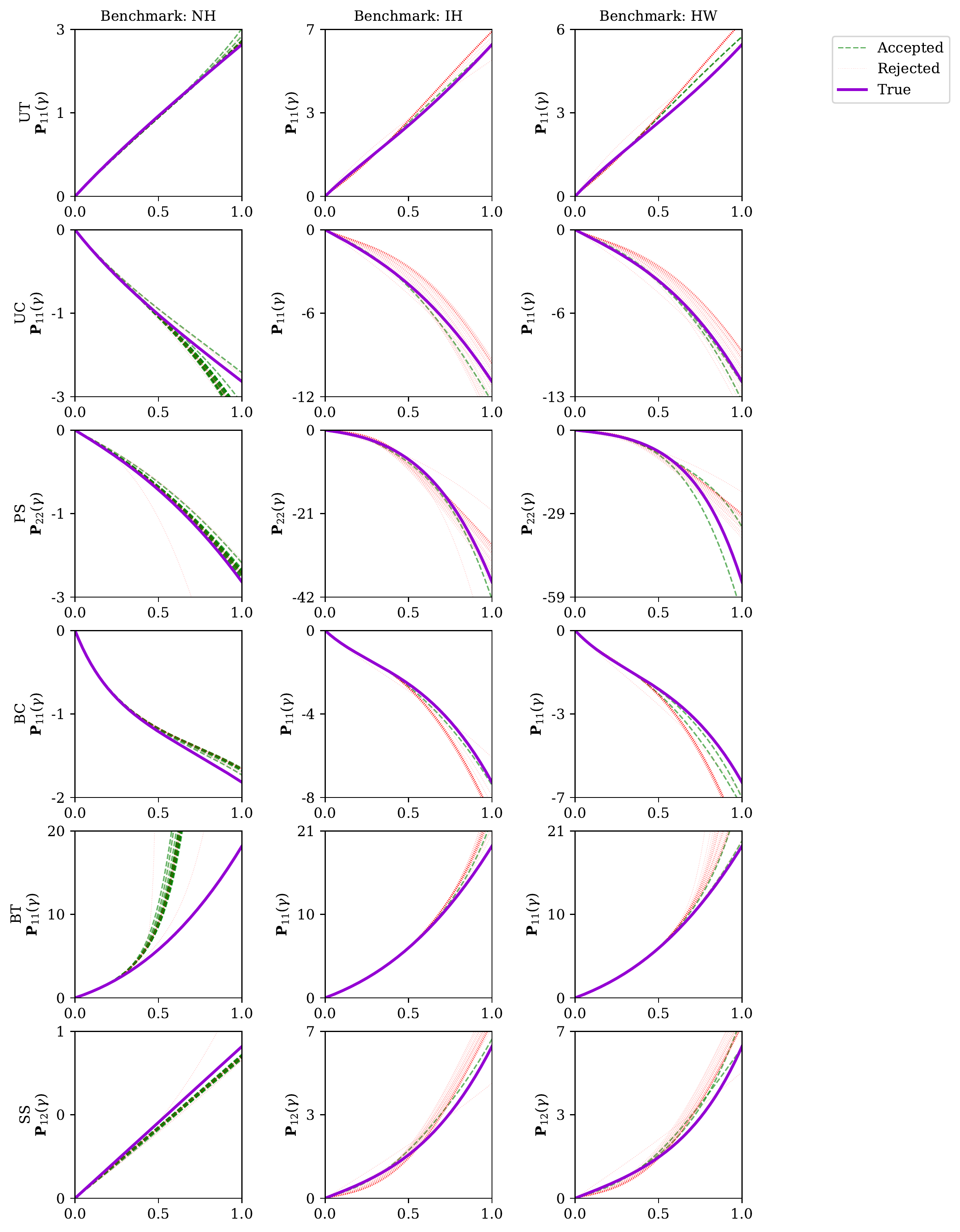}
		\caption{First Piola-Kirchhoff stress $\bfP(\bfF(\gamma))$ component prediction along the different deformation paths in \eqref{eq:strain_paths} using the ICNN-based constitutive models (both accepted and rejected ones) for the case with high noise $(\sigma_u = 10^{-3})$ and benchmarks NH \eqref{eq:NH2}, IH \eqref{eq:IH}, and HW \eqref{eq:HW}. The constitutive response of the (hidden) true model is also shown for reference.}\label{fig:NH2_IH_HW_noise=high_Pij}
\end{figure}

\begin{figure}
		\centering
		\text{Benchmark: First Piola-Kirchhoff stress component predictions, high noise ($\sigma_u=10^{-3}$)}\par\medskip
		\includegraphics[width=0.9\textwidth]{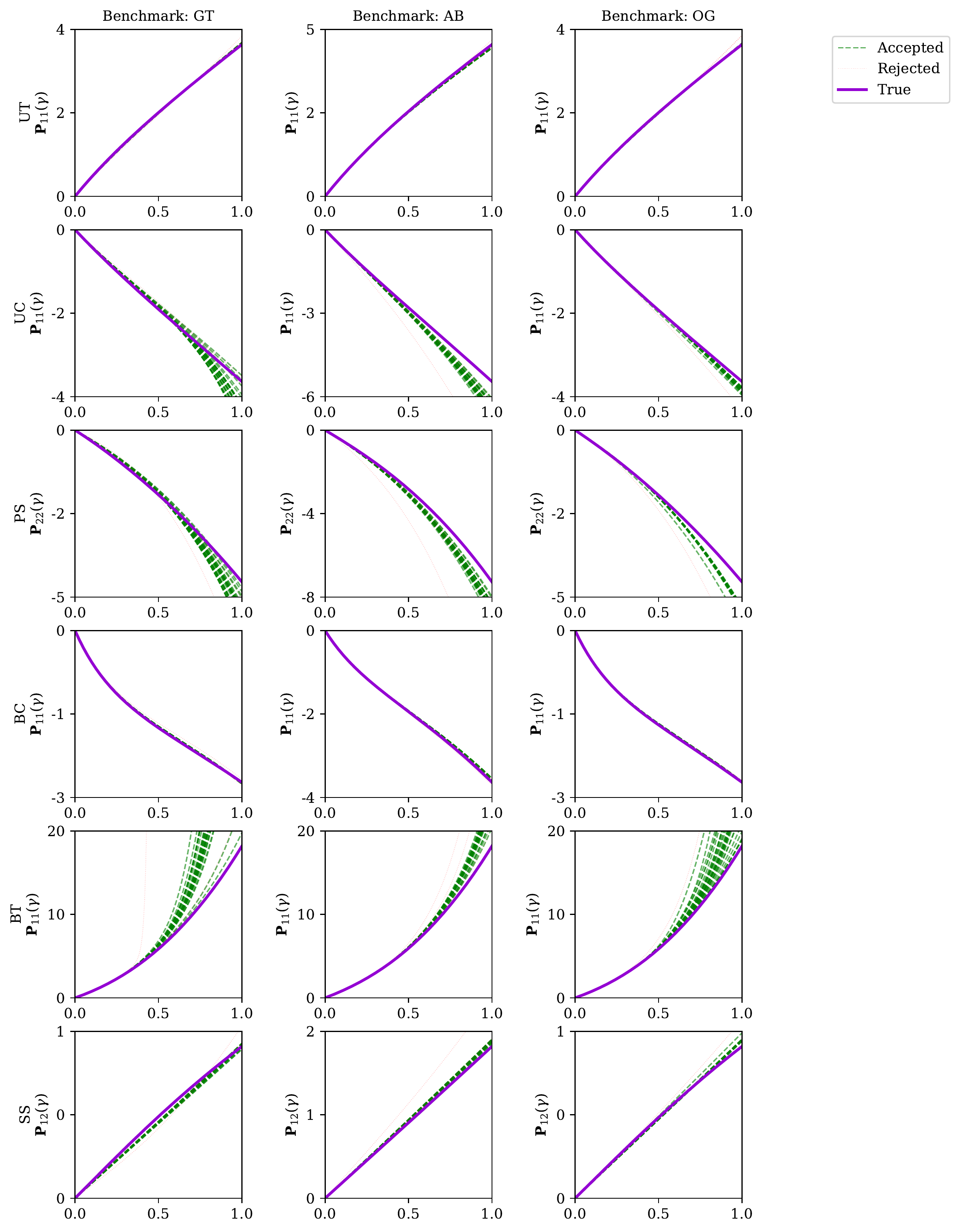}
		\caption{First Piola-Kirchhoff stress $\bfP(\bfF(\gamma))$ component prediction along the different deformation paths in \eqref{eq:strain_paths} using the ICNN-based constitutive models (both accepted and rejected ones) for the case with high noise $(\sigma_u = 10^{-3})$ and benchmarks GT \eqref{eq:GT}, AB \eqref{eq:AB}, and OG \eqref{eq:OG}. The constitutive response of the (hidden) true model is also shown for reference.}\label{fig:GT_AB_OG_noise=high_Pij}
\end{figure}

\begin{figure}
		\centering
		\text{Benchmark: First Piola-Kirchhoff stress component predictions, high noise ($\sigma_u=10^{-3}$)}\par\medskip
		\includegraphics[width=0.9\textwidth]{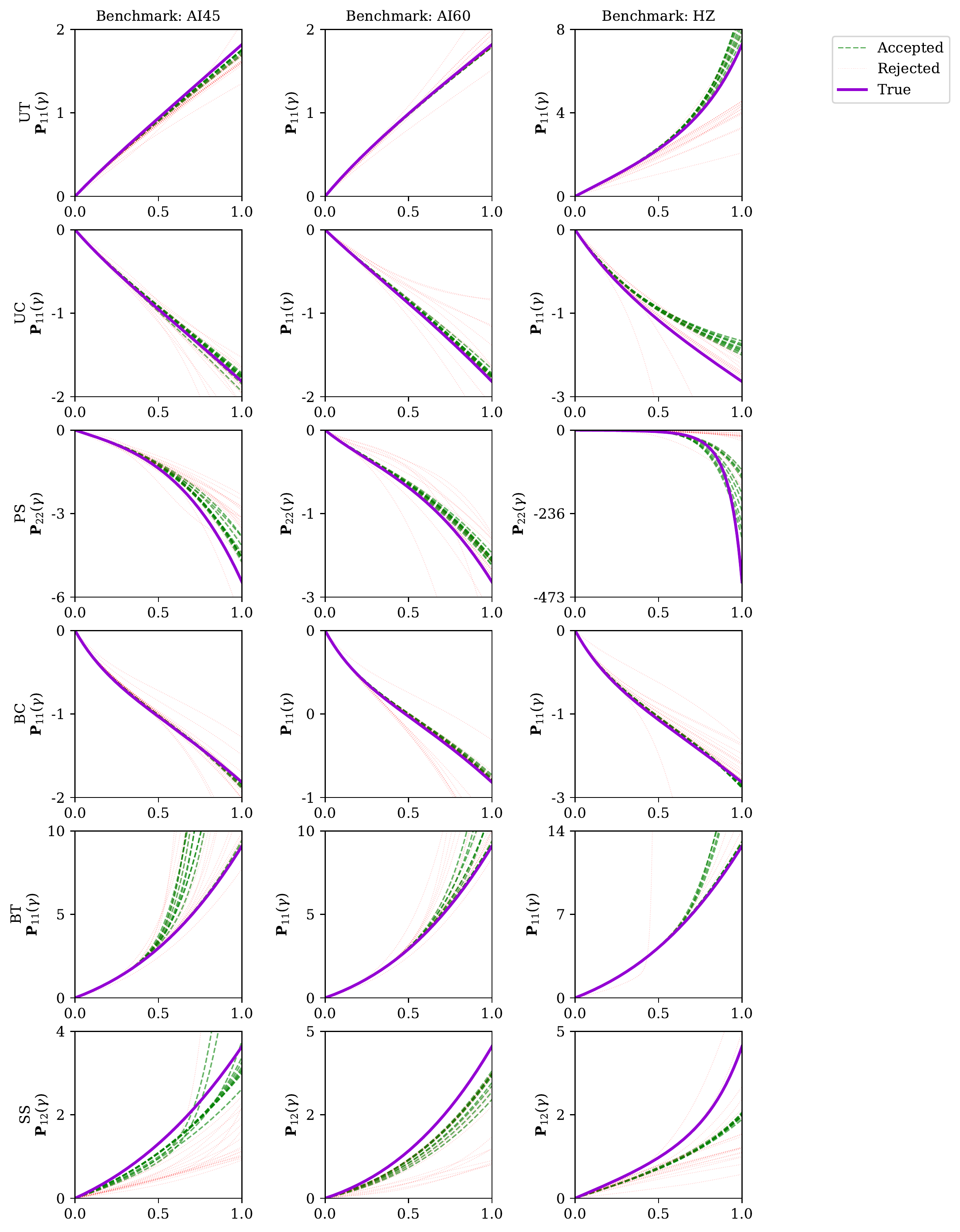}
		\caption{First Piola-Kirchhoff stress $\bfP(\bfF(\gamma))$ component prediction along the different deformation paths in \eqref{eq:strain_paths} using the ICNN-based constitutive models (both accepted and rejected ones) for the case with high noise $(\sigma_u = 10^{-3})$ and benchmarks AI45 \eqref{eq:AI45}, AI60 \eqref{eq:AI60}, and HZ \eqref{eq:HZ}. The constitutive response of the (hidden) true model is also shown for reference.}\label{fig:AI45_AI60_HA_noise=high_Pij}
\end{figure}

Figures \ref{fig:NH2_IH_HW_noise=low_Pij}-\ref{fig:AI45_AI60_HA_noise=high_Pij} show the comparisons of the first Piola-Kirchhoff stress predictions from the ICNN models vs.~the ground-truth  along the six deformation paths \eqref{eq:strain_paths} for both noise levels.



\section{Constitutive models trained without input-convex architecture and smoothness}\label{sec:non-convex-results}
\begin{figure}
		\centering
		\text{Non-ICNN benchmark: Strain-energy density predictions, high noise ($\sigma_u=10^{-3}$)}\par\medskip
		\includegraphics[width=0.9\textwidth]{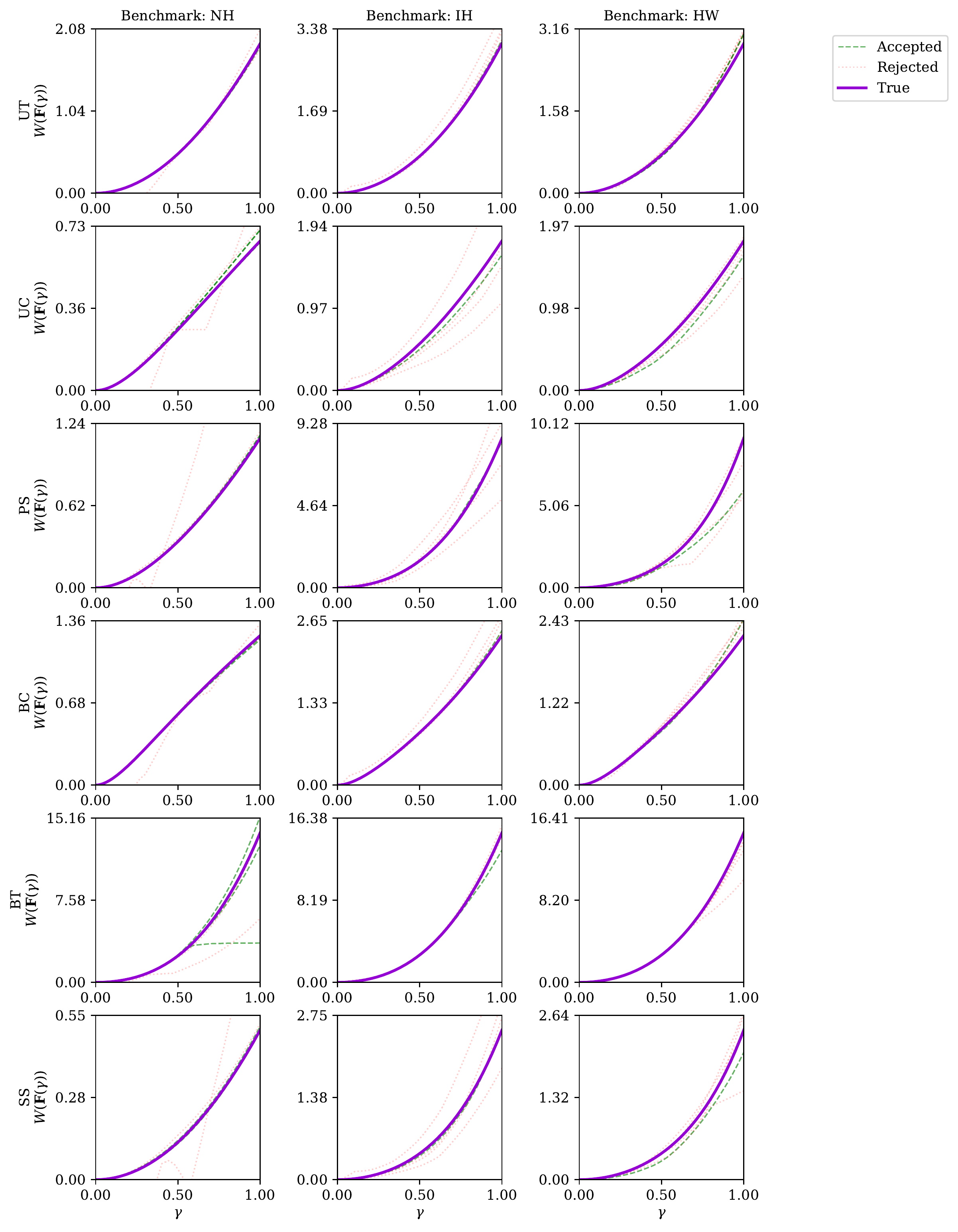}
		\caption{Strain energy density $W(\bfF(\gamma))$ prediction along the different deformation paths in \eqref{eq:strain_paths} using the NN-based (non-convex and non-smooth) constitutive models (both accepted and rejected ones) for the case with high noise $(\sigma_u = 10^{-3})$ and benchmarks NH \eqref{eq:NH2}, IH \eqref{eq:IH}, and HW \eqref{eq:HW}. The constitutive response of the (hidden) true model is also shown for reference.}\label{fig:NH_IH_HW_noise=high_W_no-convexity}
\end{figure}

\begin{figure}
		\centering
		\text{Non-ICNN benchmark: First Piola-Kirchhoff stress component predictions, high noise ($\sigma_u=10^{-3}$)}\par\medskip
		\includegraphics[width=0.9\textwidth]{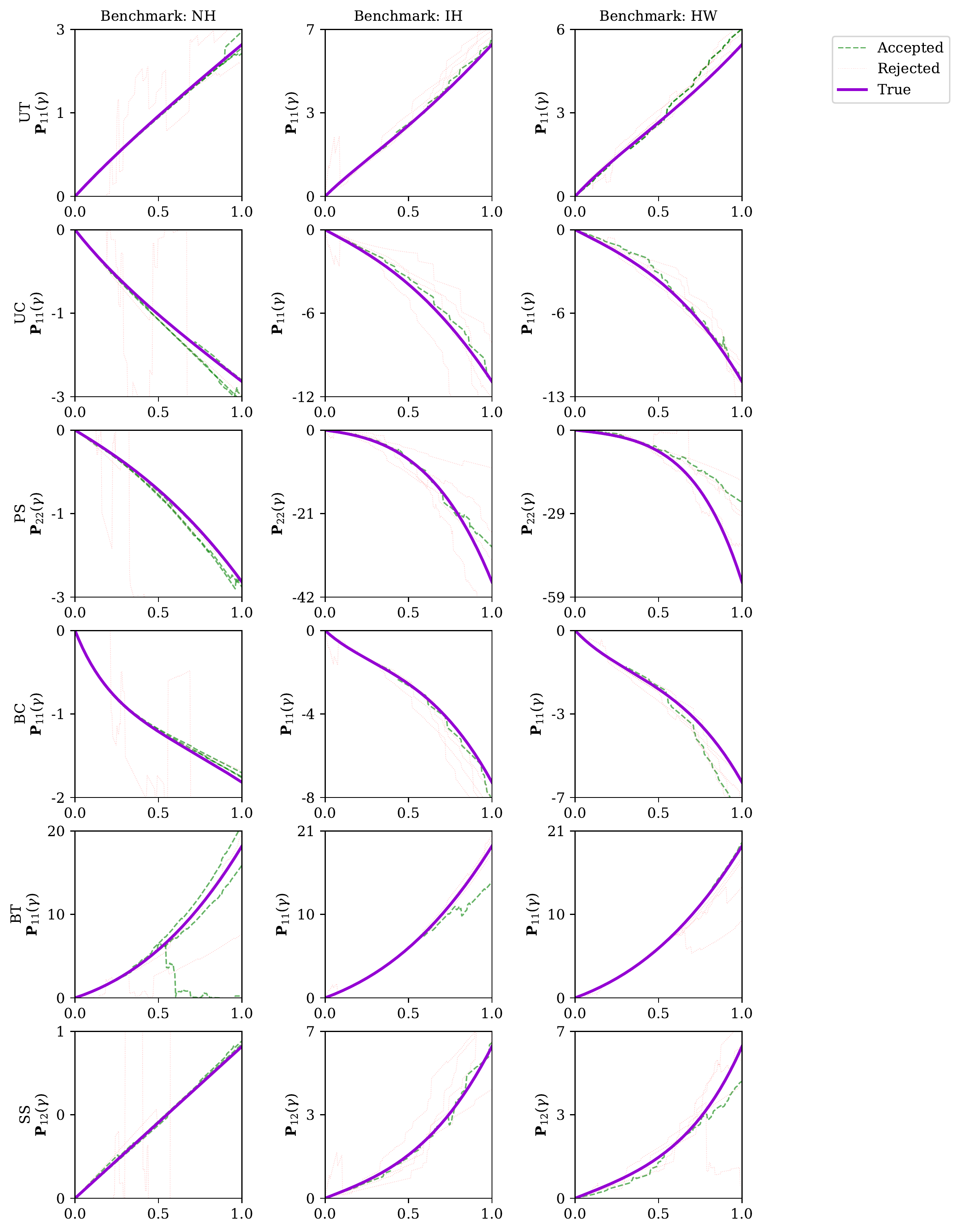}
		\caption{First Piola-Kirchhoff stress $\bfP(\bfF(\gamma))$ component prediction along the different deformation paths in \eqref{eq:strain_paths} using the NN-based (non-convex and non-smooth) constitutive models (both accepted and rejected ones) for the case with high noise $(\sigma_u = 10^{-3})$ and benchmarks NH \eqref{eq:NH2}, IH \eqref{eq:IH}, and HW \eqref{eq:HW}. The constitutive response of the (hidden) true model is also shown for reference.}\label{fig:NH_IH_HW_noise=high_Pij_no-convexity}
\end{figure}

Figures \ref{fig:NH_IH_HW_noise=high_W_no-convexity} and \ref{fig:NH_IH_HW_noise=high_Pij_no-convexity} show the constitutive response  for three representative benchmarks when the ICNN in \eqref{eq:ansatz} is replaced with a simple feed-forward neural network (keeping the same architecture)  with no constraints on the weights (i.e., no convexity constraints) and ReLU activation functions (i.e., not smooth). Although the strain energy responses are somewhat acceptable, the stress responses exhibit spurious, oscillatory, and non-smooth artifacts. Hence, a simple feed-forward NN (without enforcing smoothness and input-convexity) cannot be deployed in an FEM framework. See \citet{Asad2022} for more in-depth analysis of the ICNN architecture from a constitutive modeling perspective.

\end{document}